%% file: iclr2027_negative_transfer.tex
\documentclass[11pt]{article}
\usepackage[margin=1in]{geometry}
\usepackage[OT1]{fontenc}
\usepackage{times}
\usepackage[table]{xcolor}

\input{math_commands.tex}

\usepackage{url}
\usepackage{wrapfig}

\usepackage{booktabs}
\usepackage{tabularx}
\usepackage{multirow}
\usepackage{graphicx}
\usepackage{subcaption}
\usepackage{anyfontsize}
\usepackage{adjustbox}
\usepackage[percent]{overpic}
\usepackage{amsmath,amssymb}
\usepackage[round,authoryear]{natbib}
\usepackage[hyperfootnotes=false]{hyperref}

\newlength{\realsetupimageheight}

\definecolor{headerblue}{RGB}{220,235,248}
\definecolor{oursblue}{RGB}{240,247,252}

\hypersetup{
    colorlinks=true,
    linkcolor={[rgb]{0.10,0.24,0.48}},
    citecolor={[rgb]{0.10,0.24,0.48}},
    urlcolor={[rgb]{0.10,0.24,0.48}},
    pdfborder={0 0 0},
    pdftitle={Utility-WAM: Correcting Future-Induced Action Changes},
    pdfauthor={Feihong Zhang, Xi Lin, Yulong Shi, Zuxing Lu, Yanghong Mei,
        Zhaopeng Gu, Tingyu Yuan, Zihao Liang, Yinuo Wang, Zhaowen Li}
}

\renewenvironment{abstract}{%
    \begin{center}
        \bfseries\abstractname
    \end{center}%
    \begin{center}
        \begin{minipage}{0.90\textwidth}
            \normalsize
            \setlength{\parindent}{0pt}
            \setlength{\parskip}{0.25em}
}{%
        \end{minipage}
    \end{center}
}

\title{\fontsize{18pt}{22pt}\selectfont Utility-WAM: Correcting Future-Induced Action Changes}

\author{%
    Feihong Zhang$^{1,*}$, Xi Lin$^{2,*}$, Yulong Shi$^{3,*}$, Zuxing Lu$^{4}$, Yanghong Mei$^{5}$\\
    Zhaopeng Gu$^{5}$, Tingyu Yuan$^{5}$, Zihao Liang$^{6,\dagger}$, Yinuo Wang$^{7,\dagger}$, Zhaowen Li$^{6,\dagger}$\\[0.5ex]
    \small $^{1}$Tsinghua University\quad $^{2}$Johns Hopkins University\quad $^{3}$Northeastern University\\
    \small $^{4}$Southeast University\quad $^{5}$Institute of Automation, Chinese Academy of Sciences\\
    \small $^{6}$PAI X Lab, Yinwang Intelligent Technology Co. Ltd.\quad
    $^{7}$2030Lab, Yinwang Intelligent Technology Co. Ltd.\\[0.4ex]
    {\small $^{*}$ Equal contribution.\quad $^{\dagger}$ Corresponding authors.}
}
\date{}

\begin{document}

\maketitle

\begingroup
\renewcommand{\thefootnote}{}
\footnotetext{\raggedright $^{\dagger}$
    \mbox{\href{mailto:liangzihao1@huawei.com}{\texttt{liangzihao1@huawei.com}}};
    \mbox{\href{mailto:wangyinuo7@h-partners.com}{\texttt{wangyinuo7@h-partners.com}}};
    \mbox{\href{mailto:lizhaowen3@huawei.com}{\texttt{lizhaowen3@huawei.com}}}.\par}
\endgroup

\begin{abstract}
World Action Models (WAMs) can condition robot actions on generated visual futures, but this additional context is neither cost-free nor uniformly beneficial. We examine what a predicted future actually changes: the action delta $\Delta a$ between the future-conditioned and future-free execution paths of the same model. Under a shared WAM checkpoint, this delta barely shifts average benchmark success ($+0.8$ points on LIBERO, $-0.8$ on RoboTwin~2.0), while generating visual futures increases inference latency by 1.7--1.9$\times$. These aggregate metrics conceal a concentrated failure mode that we term \emph{future-induced negative transfer}: while predictive deltas benefit long-horizon tasks ($+2.4$ points on LIBERO-Long) and remain negligible on most routine manipulations, they induce severe degradation on precision tasks---with \emph{Press Stapler} and \emph{Place Can Basket} losing 54.0 and 34.0 points, recurring as top failure modes across independently trained WAM variants. We introduce \textbf{Utility-WAM} for in-place correction of future-conditioned control. Its primary learning component is a residual adapter that repairs future-induced velocity deltas under a frozen shared WAM by aligning them with ground-truth dynamic guidance and penalizing excess error beyond the future-free baseline. The adapter selectively eliminates the majority of severe task drops (reducing losses to 6.0 and 14.0 points) while preserving long-horizon foresight. To eliminate superfluous video computation, a lightweight pre-generation gate reuses counterfactual action outcomes to selectively bypass video generation on 75.5\% of RoboTwin queries, cutting latency by 38\% while improving average success to 93.3\% on RoboTwin~2.0 and 98.9\% on LIBERO. On physical bimanual manipulation across three real-world tasks, Utility-WAM succeeds in 53 of 60 trials, compared to 43 for FAST-WAM.
\end{abstract}

\section{Introduction}
\label{sec:introduction}

World Action Models (WAMs) couple robot control with future visual prediction, but their deployment choices differ. DreamZero~\citep{ye2026world}, Vidar~\citep{feng2507vidar}, and LingBot-VA~\citep{li2026causal} generate visual futures for action prediction; FAST-WAM~\citep{yuan2026fast} omits future video at deployment; and AdaWAM~\citep{tang2026dreaming} routes visual reasoning by execution phase. These approaches mainly ask whether or when to spend computation on visual prediction. We instead ask what a predicted future changes: the action delta $\Delta a$ it induces relative to a matched future-free path. This delta may help, hurt, or be negligible, while its video-generation cost is always paid.

FAST-WAM already suggests this tension: its separately trained future-generating variants improve LIBERO but degrade RoboTwin~2.0 relative to future-free inference~\citep{yuan2026fast}. Our controlled comparison under one shared Stage-1 checkpoint reproduces this small sign reversal (Table~\ref{tab:component_ablation}), but the averages hide where the harm lies. Following the standard definition of negative transfer as deterioration caused by an additional information source~\citep{wang2019characterizing,zhang2023survey}, we call deterioration relative to the future-free path \emph{future-induced negative transfer} (Section~\ref{sec:prelim_ntg}) and find it to be a concentrated failure mode rather than a uniform effect.

We therefore treat the action delta $\Delta a$ as the central object of study.
Empirically, this delta provides meaningful guidance on long-horizon tasks,
remains negligible on most standard manipulations (26 of 50 RoboTwin tasks change
by at most one point), but induces severe failure on precision tasks such as
\emph{Press Stapler} and \emph{Place Can Basket} (hereafter the \emph{two severe tasks}),
which also recur as the most degraded tasks across FAST-WAM variants (Appendix~\ref{app:replication}).
Because pure routing cannot rectify a corrupted action vector once future conditioning is invoked,
the foundational challenge is to \emph{repair future-induced action changes while preserving
the predictive foresight where it genuinely helps.}

To address this, we introduce \textbf{Utility-WAM}, which decouples test-time imagination into targeted action correction and selective compute allocation. Rather than discarding future generation or distilling it into a current-only policy---which compromises long-horizon performance---Utility-WAM retains the future-conditioned path while attaching a lightweight residual adapter to a frozen shared WAM. The adapter directly repairs harmful action shifts by aligning predicted velocity deltas with beneficial corrections supported by ground-truth futures. Furthermore, because physical imagination remains computationally expensive and unneeded on routine decisions, a pre-generation gate reuses the adapter's counterfactual action outcomes to selectively bypass video generation entirely when no action benefit is expected.

Our contributions are as follows:
\begin{itemize}
    \item \textbf{Characterizing Future-Induced Negative Transfer:} We formalize test-time imagination via the action delta $\Delta a$ relative to a matched future-free reference. We demonstrate that small benchmark-average effects mask severe, reproducible performance collapse on precision tasks across diverse WAM configurations.
    \item \textbf{Targeted Action Correction:} We propose a residual adapter that repairs future-induced velocity deltas via ground-truth future alignment and an excess-error penalty. On RoboTwin~2.0, the adapter recovers 89\% and 59\% of the losses on the two severe tasks (accounting for 80\% of its overall 1.7-point gain) while preserving gains on the remaining 48 tasks and long-horizon suites.
    \item \textbf{Adaptive Deployment and Physical Validation:} By deriving a pre-generation utility gate from paired adapter outcomes, Utility-WAM invokes video prediction on only 24.5\% of RoboTwin queries (cutting latency by 38\%) and achieves reported averages of 98.9\% on LIBERO and 93.3\% on RoboTwin~2.0. On real-world bimanual hardware, it improves success from 71.7\% to 88.3\% (+16.7 pp).
\end{itemize}

\section{Preliminaries}
\label{sec:preliminaries}

\paragraph{Notation and Flow-Matching WAM.}
Let $x=(o,c,s)$ denote the current observation, instruction, and
proprioceptive state, $a_{1:H}$ an action chunk, and $V$ a visual future.
The shared WAM jointly models video and action tokens with flow matching:
for a clean target $z$, noise $\epsilon$, and level $\tau$,
$z_{\tau}=(1-\tau)z+\tau\epsilon$ with target velocity $v^{*}=\epsilon-z$,
trained with the standard flow-matching MSE. Its two action paths share
parameters: $\pi_{\mathrm{wo}}$ blocks action access to future video, whereas
$\pi_{\mathrm{w}}$ conditions on its generated future.

\paragraph{Future-Induced Negative Transfer.}
\label{sec:prelim_ntg}
Treating the predicted future as the additional source of negative
transfer~\citep{wang2019characterizing}, we measure any configuration $\pi$
on evaluation episodes $\mathcal{D}$ by its success change relative to the
future-free path,
\begin{equation}
    \Delta J_{\mathcal{D}}(\pi)
    = J_{\mathcal{D}}(\pi) - J_{\mathcal{D}}(\pi_{\mathrm{wo}}),
    \label{eq:ntg}
\end{equation}
where $J_{\mathcal{D}}$ is closed-loop success. Future-induced negative
transfer is $\Delta J_{\mathcal{D}}(\pi_{\mathrm{w}})<0$; unlike classical
transfer, both policies share parameters and data. $\Delta J$ is the
closed-loop outcome of the per-query action delta $\Delta a$
(Eq.~\eqref{eq:correction_accumulation}). All tables report $\Delta J$ in
percentage points, so negative entries indicate harm.

\section{Method}
\label{sec:method}

\subsection{Overview: Core Correction and Selective Deployment}
\label{sec:method_overview}

Utility-WAM's main learning component is a residual adapter that repairs the
action changes induced by generated futures. A lightweight gate trained on the
adapter's paired outcomes then selects between this corrected path and the
future-free path. We parameterize the shared WAM, adapter,
and gate by $\theta$, $\phi$, and $\omega$, respectively.
Building on the WAM formulation in Section~\ref{sec:preliminaries}, let
$\hat v_{\mathrm{w}}$ denote the future-conditioned velocity after adapter
correction (Eq.~\eqref{eq:velocity_refinement}). At inference, the gate is
evaluated once from the current context and selects one complete action flow:
\begin{equation}
    \frac{\mathrm{d}\hat z_{\tau}}{\mathrm{d}\tau}
    =
    \begin{cases}
        \hat v_{\mathrm{w}}, & b_{\omega}(x)=1,\\
        v_{\mathrm{wo}}, & b_{\omega}(x)=0,
    \end{cases}
    \qquad
    b_{\omega}(x)=\mathbf{1}[g_{\omega}(x)>\eta],
    \label{eq:inference_routing}
\end{equation}
where $g_{\omega}(x)$ is the gate score. The future-free flow uses the
future-blocked mask; the corrected future-conditioned flow uses the
future-visible mask and applies the adapter to action velocities from
evolving video features.

For paired branches sharing initial action noise, the $N$ Euler updates
collapse into the action delta
\begin{equation}
    \Delta a
    =
    \hat a_{\mathrm{w}}-\hat a_{\mathrm{wo}}
    =
    \sum_{i=0}^{N-1}
    \Delta_i\!\left(
        \hat v_{\mathrm{w},i}-v_{\mathrm{wo},i}
    \right),
    \label{eq:correction_accumulation}
\end{equation}
where $\Delta_i$ is the $i$-th step and each velocity follows its own branch
trajectory. Both branches are run only during training to form $\Delta a$;
at inference only the selected branch runs. Stage~2 shapes the local terms,
and a binary label derived from the adapter-corrected endpoint $\Delta a$
supervises Stage~3.

\begin{figure}[t]
    \centering
    \includegraphics[width=\linewidth]{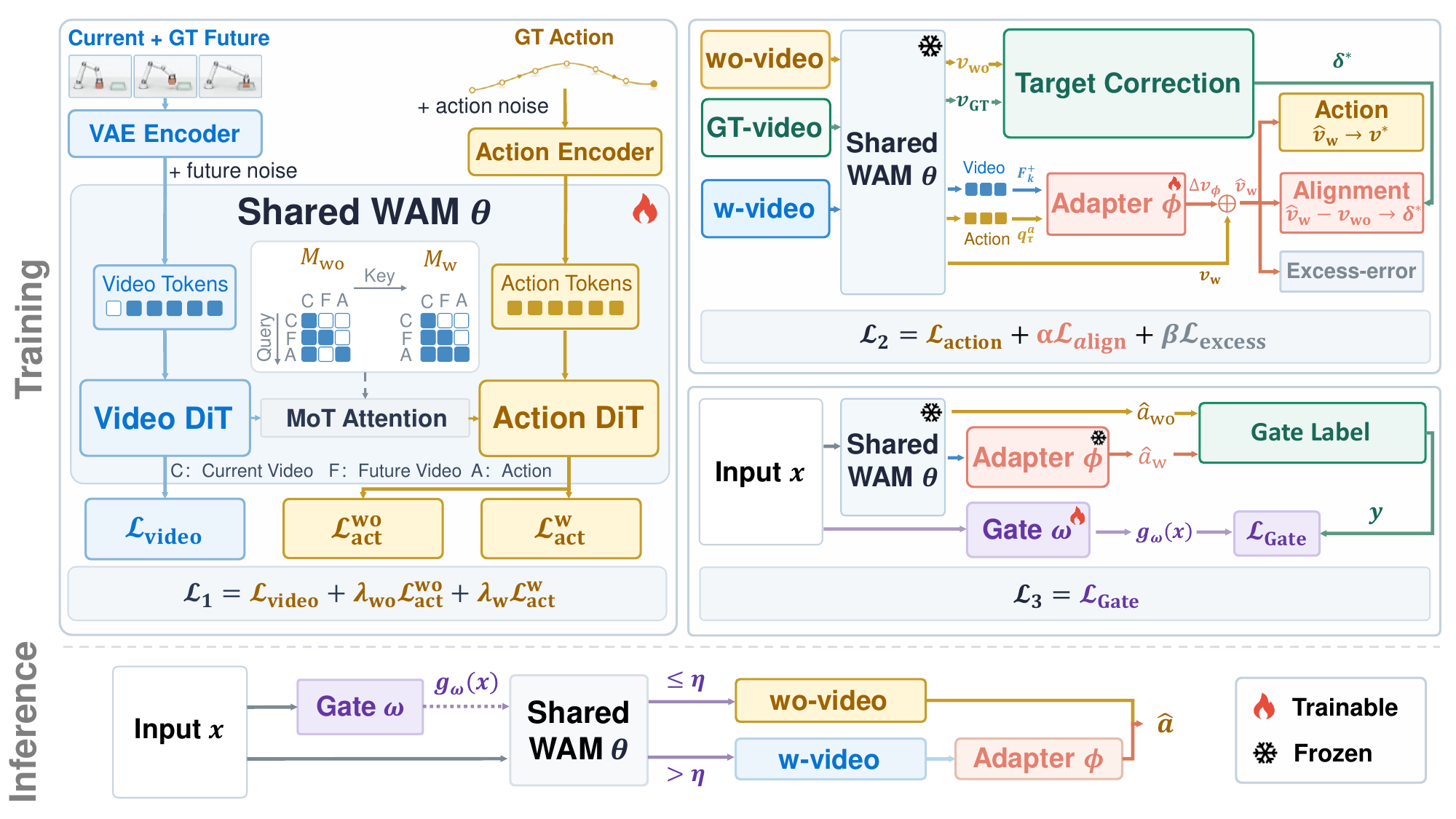}
    \caption{\textbf{Utility-WAM repairs future-induced action changes before
    allocating future-generation compute.}
    Stage~1 trains a shared WAM with matched future-free and future-visible
    action paths. Stage~2 freezes the WAM and trains a residual adapter on
    self-generated future features, using beneficial ground-truth-future
    velocity deltas and the future-free path as supervision. These two stages
    establish the core correction method. Stage~3 reuses the adapter's paired
    final-action outcomes to train a gate. At
    inference, the gate runs before video generation: it either executes the
    future-free path and skips that compute, or invokes future generation and
    the corrected future-conditioned path. Solid arrows denote inference data
    flow; dashed orange arrows denote training-only supervision.}
    \label{fig:utility_wam_concept}
\end{figure}

Training proceeds in three stages. Stage~1 jointly trains the shared video and
action DiTs under two attention masks: $M_{\mathrm{wo}}$ blocks action access
to future-video tokens, whereas $M_{\mathrm{w}}$ permits it. Both paths retain
current observations and share all parameters. We minimize
\begin{equation}
    \mathcal{L}_{1}
    =
    \mathcal{L}_{\mathrm{video}}
    +
    \lambda_{\mathrm{wo}}\mathcal{L}_{\mathrm{act}}^{\mathrm{wo}}
    +
    \lambda_{\mathrm{w}}\mathcal{L}_{\mathrm{act}}^{\mathrm{w}},
    \label{eq:stage1_loss}
\end{equation}
where the video loss and two masked action losses are balanced by
$\lambda_{\mathrm{wo}}$ and $\lambda_{\mathrm{w}}$. Stage~2 freezes $\theta$
and trains only the adapter $\phi$ on the future-conditioned path, using the
future-free velocity as its reference. Stage~3 freezes both components and
trains the gate $\omega$ from paired final-action errors. Appendix~\ref{app:detailed_training} specifies the masks and
stage-wise sampling, and Figure~\ref{fig:utility_wam_concept} summarizes the
three training stages and the resulting inference-time routing.

\subsection{Correcting Future-Induced Action Changes}
\label{sec:action_alignment}

Utility-WAM first repairs future-induced action changes. With the expert
residual $r_a^{*}=a_{1:H}-\hat a_{\mathrm{wo}}$ and $D$ valid action entries,
we measure the \emph{utility} of an action delta $\Delta a$
(Eq.~\eqref{eq:correction_accumulation}) by its paired action-error improvement:
\begin{equation}
    u_{\mathrm{act}} =
    \ell\!\left(\hat a_{\mathrm{wo}},a_{1:H}\right)
    -
    \ell\!\left(\hat a_{\mathrm{w}},a_{1:H}\right)
    =
    \frac{
        2\underbrace{\langle r_a^{*},\Delta a\rangle}_{\mathrm{alignment}}
        -
        \underbrace{\lVert\Delta a\rVert^{2}}_{\mathrm{magnitude}}
    }{D},
    \label{eq:future_utility}
\end{equation}
where $\ell$ is MSE. Negligible deltas give $u_{\mathrm{act}}\approx0$
and misdirected ones give $u_{\mathrm{act}}<0$. The adapter shapes both the
direction and the magnitude of $\Delta a$ during generation. For the same current
inputs $x$, noisy action chunk $z_{\tau}$, and noise level $\tau$, let
$v_{\mathrm{wo}}$ and $v_{\mathrm{w}}$ denote the action velocity predictions
without and with predicted future video, respectively. Their difference
$v_{\mathrm{w}}-v_{\mathrm{wo}}$ is the future-induced velocity delta; we
reserve \emph{correction} for the adapter residual introduced below.
Because $v_{\mathrm{w}}$ need not have a lower mean squared error relative to
the flow-matching target $v^{*}$ than $v_{\mathrm{wo}}$, this velocity delta is
not intrinsically beneficial.

We therefore introduce a residual adapter to correct this velocity delta. Let $q_{\tau}^{a}$ denote the action queries at action noise level $\tau$, and let $F_{k}^{+}$ denote the spatially pooled future-video features after the $k$-th video update. The adapter predicts a residual from these post-update features, yielding the refined velocity
\begin{equation}
    \hat v_{\mathrm{w}}
    =
    v_{\mathrm{w}}+\Delta v_{\phi}(q_{\tau}^{a},F_{k}^{+}).
    \label{eq:velocity_refinement}
\end{equation}

To supervise this residual, the frozen WAM receives the same noisy expert
action under no-future, predicted-future, and noised-ground-truth-future
conditioning, producing $v_{\mathrm{wo}}$, $v_{\mathrm{w}}$, and
$v_{\mathrm{GT}}$. The adapter is trained on self-generated features
$F_k^{+}$, matching its inference-time input distribution, while
$v_{\mathrm{GT}}$ identifies whether the future supports a beneficial
velocity delta. The matched-noise rollout and update sampling are detailed in
Appendix~\ref{app:detailed_training}.

Let $d(u,v)$ denote the mean squared error over valid action entries. With a relative improvement margin $\rho=0.05$, we construct
\begin{equation}
    \delta^{*}
    =
    \begin{cases}
        v_{\mathrm{GT}}-v_{\mathrm{wo}},
        & d(v_{\mathrm{GT}},v^{*})
        < (1-\rho)d(v_{\mathrm{wo}},v^{*}), \\
        0, & \text{otherwise}.
    \end{cases}
    \label{eq:target_correction}
\end{equation}
Using the refined velocity $\hat v_{\mathrm{w}}$ from Eq.~\eqref{eq:velocity_refinement}, we optimize
\begin{equation}
    \mathcal{L}_{2} = \mathbb{E}\!\left[\underbrace{w(\tau)\,d(\hat v_{\mathrm{w}},v^{*})}_{\text{Action}} + \alpha\underbrace{d(\hat v_{\mathrm{w}}-v_{\mathrm{wo}},\delta^{*})}_{\text{Alignment}} + \beta\underbrace{\max\!\left\{d(\hat v_{\mathrm{w}},v^{*})-d(v_{\mathrm{wo}},v^{*}),\,0\right\}}_{\text{Excess-error}}\right].
    \label{eq:stage2_loss}
\end{equation}
Here $w(\tau)$ is the FAST-WAM scheduler weight, and $\alpha,\beta$ weight
alignment and excess error. The three objective terms jointly regulate the direction and magnitude of the action correction: the action term fits expert demonstration velocities, the alignment term steers $\hat v_{\mathrm{w}}-v_{\mathrm{wo}}$ toward the velocity shift supported by ground-truth futures (collapsing to zero when privileged futures offer no advantage), and the excess-error regularizer penalizes local flow-matching error exceeding the future-free baseline.

\paragraph{Why correction is applied to the future-conditioned path.}
Both execution paths share identical current observations, but the future-conditioned branch incorporates synthetic representations generated through noisy sampling. While test-time visual imagination provides dynamic foresight on complex horizons, imperfect video generation introduces distribution shifts that degrade action precision. Equation~\eqref{eq:stage2_loss} leverages $v_{\mathrm{wo}}$ as an anchor: when ground-truth future conditioning yields negligible benefit, the adapter suppresses deviations from the reliable future-free reference; when future conditioning is demonstrably beneficial, it mirrors this correction in the spirit of professor forcing~\citep{lamb2016professor}. Unlike PFD, which distills future-induced velocity residuals into a current-only policy~\citep{fang2026privileged}, this preserves visual foresight while constraining its negative side effects.

\subsection{Skipping Futures that Correction Does Not Make Useful}
\label{sec:utility_gating}

Correction is soft: the excess-error term constrains local velocity error but
does not make every endpoint correction useful, and the corrected branch still
generates video on every query. The adapter's paired outcomes identify where
this happens. Let $y=\mathbf{1}[u_{\mathrm{act}}>0]$, with $u_{\mathrm{act}}$
(Eq.~\eqref{eq:future_utility}) evaluated on the adapter-corrected $\Delta a$.
Because $\Delta a$ is unavailable before video generation, a lightweight gate
learns $g_{\omega}(x)\approx P(y=1\mid x)$ from current inputs with
$\mathcal{L}_{3}=\mathcal{L}_{\mathrm{Gate}}=\mathrm{BCE}(g_{\omega}(x),y)$
(Appendix~\ref{app:detailed_training}), so queries on which correction yields
no improvement run the future-free path and skip video generation. Sweeping
its threshold traces the accuracy--compute frontier in
Figure~\ref{fig:neff_success_curves}; Appendix~\ref{app:threshold_selection}
gives the budgeted routing objective, default calibration, and hard-budget
rule.

\subsection{Inference}
\label{sec:inference}

We evaluate the gate before video generation and execute only the selected
path. The future-free path performs ten action updates; the future-conditioned
path pairs each video update with an adapter-refined action update. We fix
$\eta$ from held-out validation data before test evaluation. Appendices~\ref{app:detailed_training}
and~\ref{app:threshold_selection} give
the solver, update order, and threshold-selection rule.

\section{Experiments}
\label{sec:experiments}
Experiments first test the core prediction: correction should shrink the severe
losses while retaining performance on the remaining tasks and the long-horizon
gain. We then test whether the adapter-derived gate improves on always-on
correction at lower compute and against compute-matched baselines.

\subsection{Simulation Experiments}
\label{sec:simulation_experiments}

\paragraph{Setup.}
We build on FAST-WAM~\citep{yuan2026fast} with a Wan2.2-TI2V-5B backbone
and train one model per benchmark. All controlled variants share the same
Stage-1 checkpoint, isolating the effects of adaptation and routing, and are
evaluated on identical initial states and random seeds. LIBERO evaluation covers 40 tasks and 2,000 trials, executing 10 of
32 predicted actions; RoboTwin~2.0 uses 100 trials per task under clean and
randomized conditions, executing 24 of 32 actions. Appendix~\ref{app:model_and_hyperparameters} and
Appendix~\ref{app:sim_bench} provide architecture, optimization, data, and
rollout details.

\subsection{Diagnosing Future-Induced Negative Transfer}
\label{sec:ablation_efficiency}

\paragraph{Benchmark Level: A Small Average Effect.}
Relative to Shared WAM (wo), future conditioning changes average success by
$+0.8$ points on LIBERO and $-0.8$ points on RoboTwin~2.0
(Table~\ref{tab:component_ablation}). The LIBERO gain is concentrated on
Long ($+2.4$ points versus at most $+0.4$ elsewhere), and the RoboTwin
decrease is not larger under randomization ($-1.1$ Clean, $-0.5$ Random).
FAST-WAM's separately trained Joint and IDM variants show the same trend
(Appendix~\ref{app:replication}).

\paragraph{Task Level: Concentrated Severe Failures.}
The averages hide where the harm lies. Table~\ref{tab:ntg_task_summary}
summarizes the task-level changes $\Delta J_{t}(\pi_{\mathrm{w}})$. On
RoboTwin~2.0, 21 of 50 tasks have $\Delta J_{t}<0$, yet the median is 0.0:
the aggregate loss is driven by \emph{Press Stapler} ($-54.0$) and
\emph{Place Can Basket} ($-34.0$), while the other 19 harmed tasks lose at
most 2.5 points each. These two tasks are also among the three most harmed
for both FAST-WAM future variants
(Appendix Table~\ref{tab:robotwin_per_task_ntg}), so the severe failures
recur across three future-usage mechanisms. We therefore treat them as the
primary targets for correction; Appendix~\ref{app:task_level_ntg} relates
each task's loss to its adapter gain.

\begin{table}[htbp]
      \centering
      \caption{\textbf{Task-level prevalence and severity of future-induced
      negative transfer.} Entries are $\Delta J$ of Shared WAM (w) in
      percentage points (negative values indicate harm). ``Harmed'' counts
      tasks with $\Delta J_{t}<0$. Entries are computed from the complete
      per-task simulation evaluation logs; Appendix
      Table~\ref{tab:robotwin_per_task_ntg} lists the ten most harmed
      RoboTwin~2.0 tasks.}
      \label{tab:ntg_task_summary}
      \small
      \setlength{\tabcolsep}{4pt}
      \begin{adjustbox}{max width=\linewidth}
      \begin{tabular}{lcccc}
          \toprule
          Evaluation & Aggregate $\Delta J$ & Harmed tasks
          & Median $\Delta J_{t}$ & $\Delta J_{t}$ range \\
          \midrule
          LIBERO
          & $+0.8$
          & $7/40$
          & $0.0$
          & $[-6.0,+6.0]$ \\

          RoboTwin~2.0
          & $-0.8$
          & $21/50$
          & $0.0$
          & $[-54.0,+5.5]$ \\

          \bottomrule
      \end{tabular}
      \end{adjustbox}
  \end{table}

\begin{table}[htbp]
    \centering
    \setlength{\belowcaptionskip}{6pt}
    \caption{\textbf{Controlled diagnosis and mitigation of future-induced negative transfer.}
    All entries are success rates (\%) under a shared Stage-1 checkpoint.
    The fixed w-versus-wo comparison measures the aggregate effect of future
    conditioning. Utility-WAM (Correction) isolates the core correction
    method with future generation always enabled; Utility-WAM (Selective)
    adds the adapter-derived gate.
    Bold and underlined entries denote the best and second-best
    results in each column, respectively (including ties).}
    \label{tab:component_ablation}

    \normalsize 
    \setlength{\tabcolsep}{4pt}
    \renewcommand{\arraystretch}{1.15}

    \begin{adjustbox}{max width=\linewidth}
        \begin{tabular}{lcccccccc}
            \toprule
            \multirow{2}{*}{Method}
            & \multicolumn{5}{c}{LIBERO}
            & \multicolumn{3}{c}{RoboTwin 2.0} \\
            \cmidrule(lr){2-6}\cmidrule(l){7-9}
            & Spatial & Object & Goal & Long & Avg.
            & Clean & Random & Avg. \\
            \midrule

            Shared WAM (wo)
            & 97.2 & 99.0 & \underline{98.0} & 95.4 & 97.4
            & 92.4 & 91.3 & 91.9 \\

            Shared WAM (w)
            & 97.6 & \underline{99.2} & \underline{98.0}
            & 97.8 & 98.2
            & 91.3 & 90.8 & 91.1 \\

            \midrule
            Utility-WAM (Correction)
            & \textbf{98.6} & \textbf{100.0} & 97.8
            & \underline{98.4} & \underline{98.7}
            & \underline{93.3} & \underline{92.3}
            & \underline{92.8} \\

            Utility-WAM (Selective)
            & \underline{98.4} & \textbf{100.0} & \textbf{98.2}
            & \textbf{98.8} & \textbf{98.9}
            & \textbf{93.8} & \textbf{92.7} & \textbf{93.3} \\

            \bottomrule
        \end{tabular}
    \end{adjustbox}
\end{table}

\subsection{Core Correction and Selective Deployment}
\label{sec:mitigation}

We evaluate mitigation in training order (Tables~\ref{tab:component_ablation} and~\ref{tab:adapter_loss_ablation}).

\begin{table}[t]
    \centering
    \setlength{\belowcaptionskip}{4pt}
    \caption{\textbf{Adapter-objective ablations and selective routing.}
    Success entries are benchmark averages (\%) under a shared Stage-1
    checkpoint. All rows share identical evaluation episodes and random seeds.}
    \label{tab:adapter_loss_ablation}
    \small
    \setlength{\tabcolsep}{3.5pt}
    \renewcommand{\arraystretch}{1.06}
    \begin{adjustbox}{max width=\linewidth}
    \begin{tabular}{lllccc}
        \toprule
        Configuration & Routing & Adapter objective
        & LIBERO & RoboTwin 2.0
        & \shortstack{$N_{\mathrm{eff}}$\\LIB./RT} \\
        \midrule
        Shared WAM (wo) & Fixed: no video & -- & 97.4 & 91.9 & 0/0 \\
        Shared WAM (w) & Fixed: video & -- & 98.2 & 91.1 & 10/10 \\
        \midrule
        Action only & Fixed: video & Action & 98.2 & 91.4 & 10/10 \\
        + Alignment & Fixed: video & Action + Align. & 98.5 & 92.3 & 10/10 \\
        + Excess-error & Fixed: video & Action + Excess & 98.3 & 91.8 & 10/10 \\
        Utility-WAM (Correction) & Fixed: video & Full & \underline{98.7}
        & \underline{92.8} & 10/10 \\
        \midrule
        \rowcolor{oursblue}
        Utility-WAM (Selective) & Learned & Full & \textbf{98.9} & \textbf{93.3}
        & 8.70/2.45 \\
        \bottomrule
    \end{tabular}
    \end{adjustbox}
\end{table}

\paragraph{Core Result: Gains Concentrate Where Futures Hurt, Without Sacrificing Other Tasks.}
The adapter-loss rows in Table~\ref{tab:adapter_loss_ablation} isolate the objective with the shared WAM frozen, future generation enabled, and the gate disabled. Training the adapter on self-generated futures with action loss alone ($\mathcal{L}_{\text{action}}$), akin to scheduled sampling~\citep{bengio2015scheduled}, changes success by $\le 0.3$ points; this capacity-matched control confirms that added parameters alone cannot resolve future-induced degradation.

Instead, the gains stem from the complementary pairing of alignment and error regularization in Eq.~\eqref{eq:stage2_loss}. The alignment term ($\mathcal{L}_{\text{align}}$) provides constructive steering toward privileged corrections $\delta^*$, recovering +0.9 points on RoboTwin~2.0 and +0.3 on LIBERO. Meanwhile, the excess-error penalty ($\mathcal{L}_{\text{excess}}$) serves as an asymmetric safety anchor: it remains largely inactive on routine tasks, but penalizes velocity drift exceeding the future-free baseline. Together, they curb residual degradation during noisy rollouts, contributing an essential +0.5 points on RoboTwin~2.0 (reaching 92.8\%) and +0.2 on LIBERO (98.7\%).

Under always-video routing, this combined objective produces the intended targeted-recovery pattern: gains concentrate where futures cause severe harm. The two severe tasks supply 1.36 of its 1.7-point RoboTwin gain (80\%): their losses shrink from 54.0 to 6.0 and from 34.0 to 14.0 points, where the excess-error penalty crucially arrests catastrophic drift. The other 48 tasks still improve by $+0.35$ points on average, and LIBERO-Long keeps its gain (98.4\% versus 97.8\%). Thus, correction removes concentrated harm without sacrificing the benefit of futures (Appendix Table~\ref{tab:robotwin_per_task_ntg}); LIBERO improves by $0.5$ points.

\paragraph{Gate: Skipping Futures that Correction Does Not Make Useful.}

\begin{figure}[htbp]
    \centering
    \includegraphics[width=\linewidth]{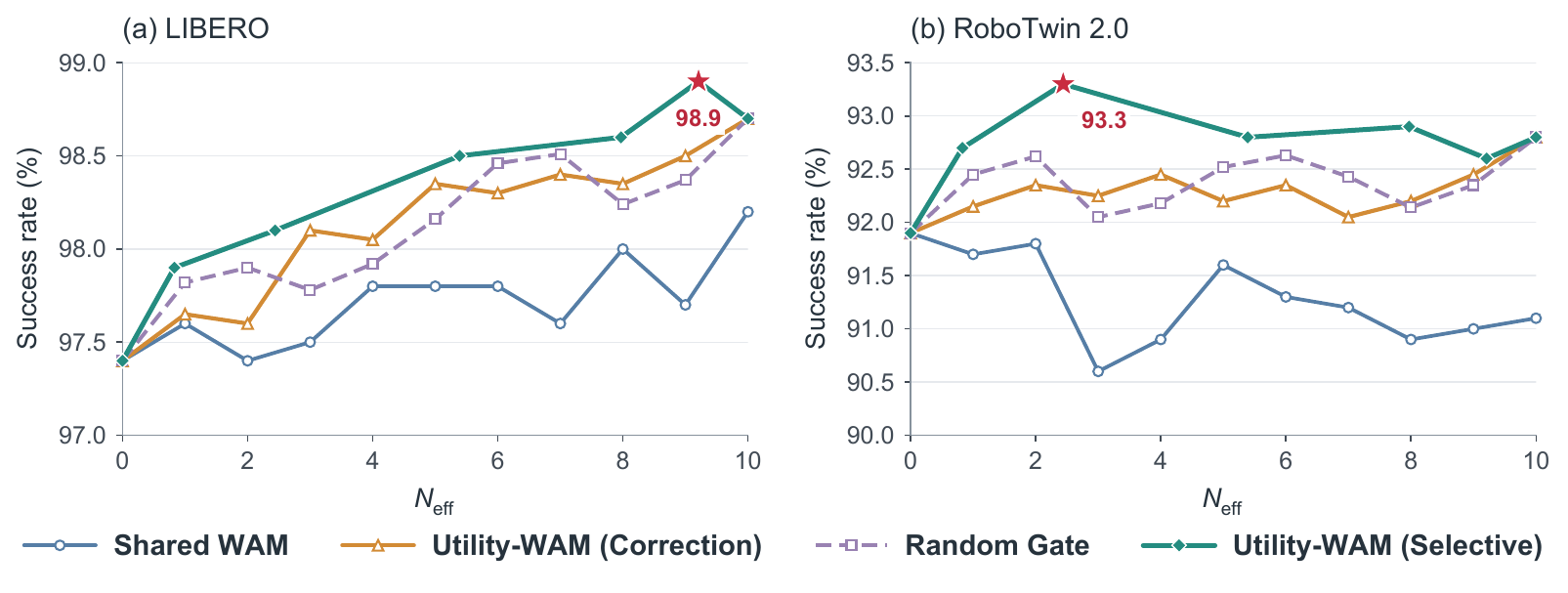}
    \caption{\textbf{Accuracy--compute frontier of selective routing.}
        Success rate versus effective video generation steps on
        LIBERO and RoboTwin 2.0; moving left saves video generation.
        $N_{\mathrm{eff}}$ denotes the average number of video generation
        steps per action query.
        Red stars mark the validation-calibrated operating points; the
        remaining points form a test-time threshold sweep.
        The selective configuration is compared with two compute-matched baselines. The
        correction-only curve is a static schedule that updates the future video
        only during the first $n$ of the ten denoising steps of every query,
        so that $N_{\mathrm{eff}}=n$. The dashed Random Gate curve routes each
        query to the corrected future-conditioned path with probability
        $N_{\mathrm{eff}}/10$.
    }
    \label{fig:neff_success_curves}
\end{figure}

We fix the gate thresholds on the validation sets ($\eta=0.439$ for LIBERO and $0.541$ for RoboTwin~2.0; Appendix~\ref{app:threshold_selection}). Figure~\ref{fig:neff_success_curves} reports success against $N_{\mathrm{eff}}$ as we sweep the threshold, tracing the accuracy--compute frontier of Eq.~\eqref{eq:budgeted_routing}.

At the calibrated operating points, the gate generates video on 87\% of
LIBERO queries and 24.5\% of RoboTwin~2.0 queries
($N_{\mathrm{eff}}=8.70$ and $2.45$) and raises success from 98.7\% and
92.8\% under always-on correction to 98.9\% and 93.3\%. The gain comes from
queries that correction does not make useful: on the 48 RoboTwin~2.0 tasks
outside the two severe ones, the gain over the future-free path rises from
$+1.35$ to about $+1.9$ points, whereas the two severe tasks keep the recovery
obtained by correction (Appendix~\ref{app:task_level_ntg}). On RoboTwin~2.0
the gate cuts latency by 38\% relative to always-video inference, adding
17\% over the future-free path for $+1.4$ points; on LIBERO, where futures
help on average (Section~\ref{sec:ablation_efficiency}), it keeps most
queries on the future-conditioned path and saves 6.7\%
(Appendix Table~\ref{tab:inference_efficiency}).

On both benchmarks, the gate curve lies above adapter-only static truncation and random routing at every evaluated intermediate budget ($0<N_{\mathrm{eff}}<10$). Random routing reaches at most 98.4\% on LIBERO ($N_{\mathrm{eff}}\in\{8,9\}$) and 92.6\% on RoboTwin~2.0 ($N_{\mathrm{eff}}\in\{2,3\}$), below the gate at its operating points, and truncating video generation on every query also remains below the gate. The learned gate therefore improves the measured accuracy--compute frontier over these two compute-matched controls. Unlike phase-supervised routing~\citep{tang2026dreaming}, the gate needs no phase labels and applies unchanged to variable-length real-world episodes.

\subsection{Comparison with Reported Results}
\label{sec:reported_results}

Table~\ref{tab:main_results} compares Utility-WAM with representative VLAs
and WAMs, with FAST-WAM as the primary reported comparison because we follow
its data and evaluation pipeline. Its future-generating Joint and IDM variants
raise LIBERO-Long from 95.2\% to 96.8\% and 97.6\%, respectively, but lower
RoboTwin~2.0 from 91.8\% to 90.6\% and 91.3\%. The adapter alone already
exceeds every reported LIBERO average (98.7\%) and is within 0.1 points of
the best reported RoboTwin~2.0 average (92.8\% versus 92.9\%); with the gate, Utility-WAM reaches
98.9\% and 93.3\%, retaining the long-horizon benefit of generated futures
while removing their concentrated harm. While cross-paper comparisons involve independently trained models, the controlled ablations in Table~\ref{tab:component_ablation} isolate the exact contribution of each component under a shared backbone.

\begin{table}[htbp]
    \centering
    \caption{\textbf{Comparison with reported results} on LIBERO and
    RoboTwin~2.0. $\dagger$ marks values not reported in the cited papers. Bold/underline: best/second-best (including
    ties). Sources are in Appendix~\ref{app:baseline_protocols}.}
    \label{tab:main_results}
    \small
    \setlength{\tabcolsep}{2.5pt}
    \renewcommand{\arraystretch}{1.08}
    \begin{adjustbox}{max width=\linewidth}
    \begin{tabular}{lcccccccc}
        \toprule
        & \multicolumn{5}{c}{LIBERO}
        & \multicolumn{3}{c}{RoboTwin 2.0} \\
        \cmidrule(lr){2-6}\cmidrule(l){7-9}
        Method
        & Spatial & Object & Goal & Long & Avg.
        & Clean & Random & Avg. \\
        \midrule
        $\pi_{0.5}$\textsuperscript{$\dagger$}~\citep{intelligence2025pi_}
        & \underline{98.8} & 98.2 & 98.0 & 92.4 & 96.9
        & 82.7 & 76.8 & 79.8 \\
        LingBot-VA~\citep{li2026causal}
        & 98.5 & \underline{99.6} & 97.2 & 98.5 & 98.5
        & 92.9 & 91.6 & 92.2 \\
        Cosmos-Policy~\citep{kim2026cosmos}
        & 98.1 & \textbf{100.0} & \underline{98.2} & 97.6 & 98.5
        & -- & -- & -- \\
        Efficient-WAM~\citep{li2026efficient}
        & -- & -- & -- & -- & -- & 86.7 & 85.7 & 86.2 \\
        AHA-WAM~\citep{cai2026aha}
        & -- & -- & -- & -- & -- & \underline{93.4} & 92.2
        & 92.8 \\
        AdaWAM~\citep{tang2026dreaming}
        & 98.0 & \underline{99.6} & 97.1 & \textbf{99.1} & 98.5
        & 93.1 & 91.4 & 92.2 \\
        \midrule
        FAST-WAM~\citep{yuan2026fast}
        & 98.2 & \textbf{100.0} & 97.0 & 95.2 & 97.6
        & 91.9 & 91.8 & 91.8 \\
        FAST-WAM-Joint~\citep{yuan2026fast}
        & \textbf{99.6} & 99.4 & \underline{98.2} & 96.8 & 98.5
        & 90.8 & 90.3 & 90.6 \\
        FAST-WAM-IDM~\citep{yuan2026fast}
        & \underline{98.8} & 97.8 & 97.8 & 97.6 & 98.0
        & 91.2 & 91.3 & 91.3 \\
        PFD~\citep{fang2026privileged}
        & 98.6 & 99.2 & \textbf{99.2} & 95.4 & 98.1
        & 93.1 & \textbf{92.7} & \underline{92.9} \\
        \midrule
        Utility-WAM (Correction)
        & 98.6 & \textbf{100.0} & 97.8 & 98.4 & \underline{98.7}
        & 93.3 & \underline{92.3} & 92.8 \\
        \rowcolor{oursblue}
        \textbf{Utility-WAM (Selective)}
        & 98.4 & \textbf{100.0} & \underline{98.2} & \underline{98.8}
        & \textbf{98.9} & \textbf{93.8} & \textbf{92.7} & \textbf{93.3} \\
        \bottomrule
    \end{tabular}
    \end{adjustbox}
\end{table}

\subsection{Real-World Evaluation}
\label{sec:real_world}

\begin{figure}[t]
    \centering
    \setlength{\realsetupimageheight}{0.13\linewidth}
    \captionsetup[subfigure]{
        font=footnotesize,
        labelfont=bf,
        justification=centering,
        skip=2pt
    }
    \begin{minipage}[c]{0.50\linewidth}
    \centering
    \begin{subfigure}[t]{0.49\linewidth}
        \vspace{0pt}
        \centering
        \begin{minipage}[c][\realsetupimageheight][c]{\linewidth}
            \centering
            \includegraphics[
                width=\linewidth,
                height=\realsetupimageheight,
                keepaspectratio
            ]{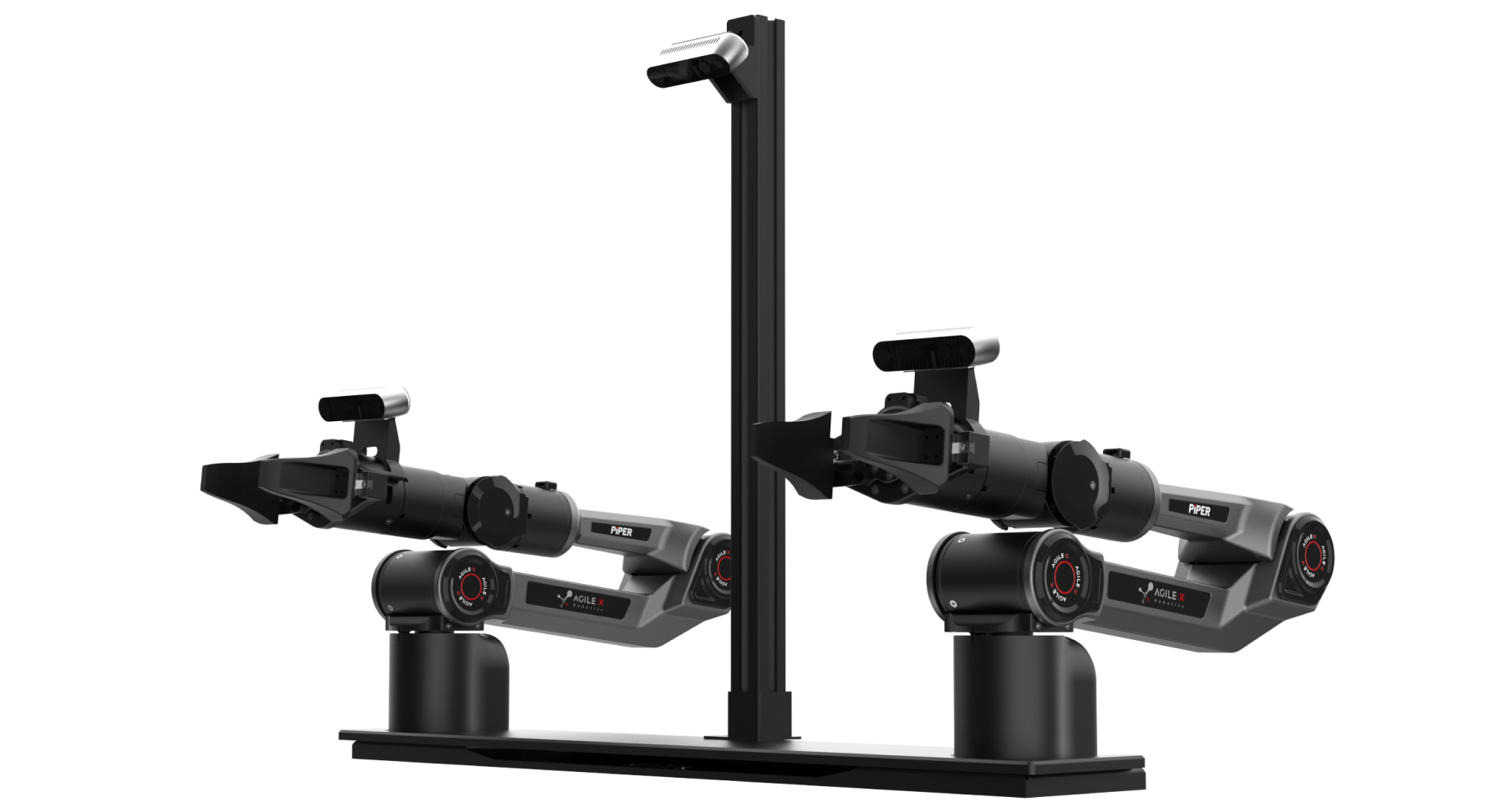}
        \end{minipage}
        \caption{AgileX PiPER}
        \label{fig:real_platform}
    \end{subfigure}\hfill
    \begin{subfigure}[t]{0.49\linewidth}
        \vspace{0pt}
        \centering
        \adjincludegraphics[
            trim={{0.215\width} {0.014\height}
                  {0.410\width} {0.486\height}},
            clip,
            width=\linewidth,
            height=\realsetupimageheight,
            keepaspectratio
        ]{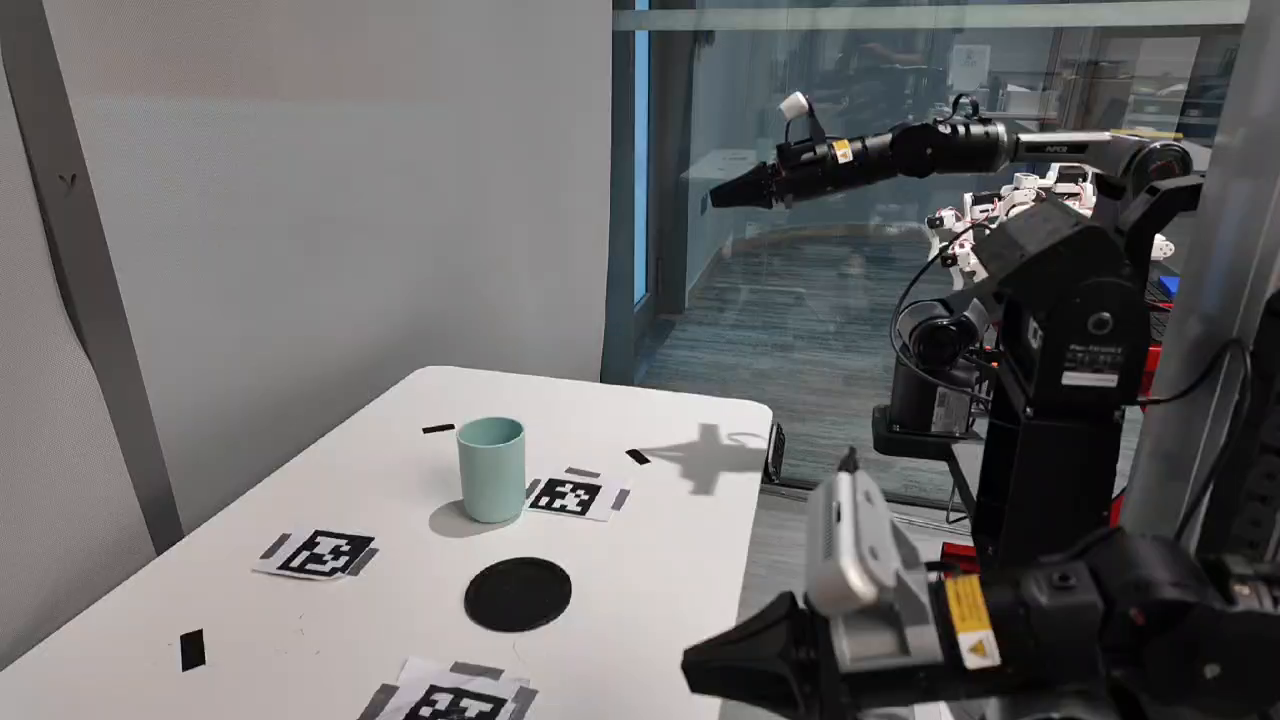}
        \caption{Place Empty Cup}
        \label{fig:real_cup}
    \end{subfigure}\\[2pt]
    \begin{subfigure}[t]{0.49\linewidth}
        \vspace{0pt}
        \centering
        \adjincludegraphics[
            trim={{0.40625\width} {0.31250\height}
                  {0.25000\width} {0.22917\height}},
            clip,
            width=\linewidth,
            height=\realsetupimageheight,
            keepaspectratio
        ]{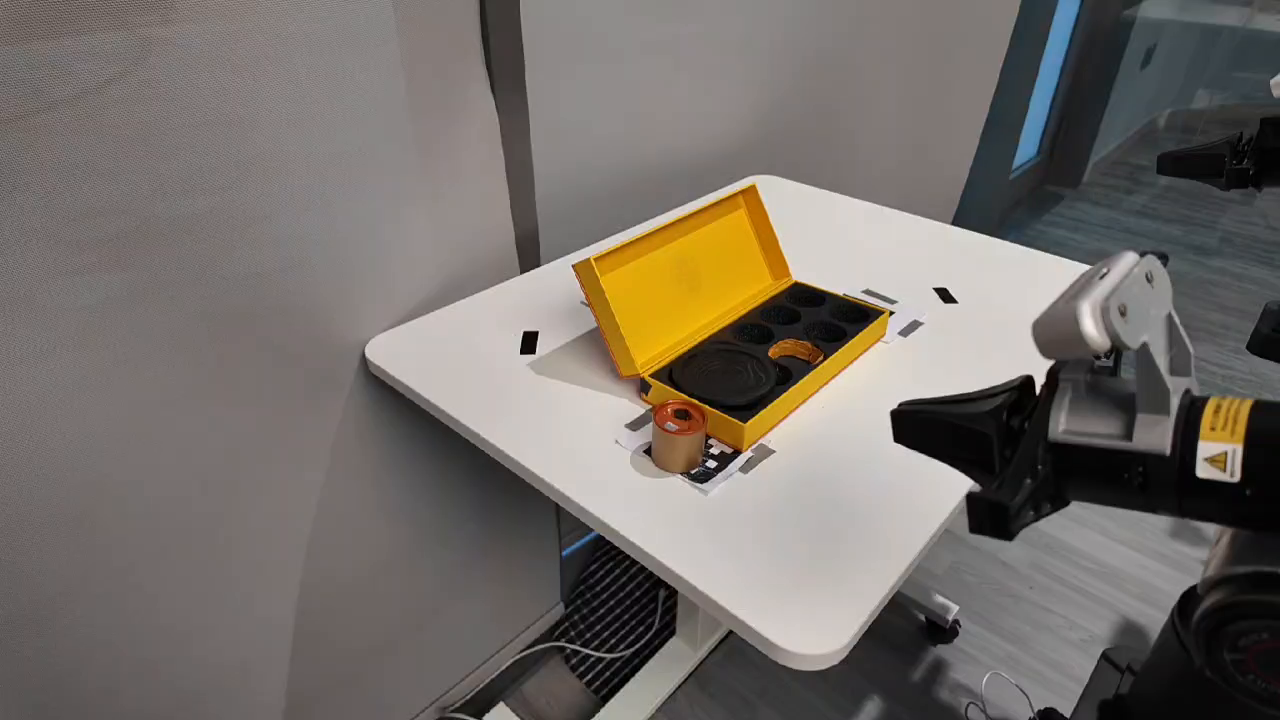}
        \caption{Insert Can}
        \label{fig:real_can}
    \end{subfigure}\hfill
    \begin{subfigure}[t]{0.49\linewidth}
        \vspace{0pt}
        \centering
        \adjincludegraphics[
            trim={{0.234375\width} {0.083333\height}
                  {0.390625\width} {0.416667\height}},
            clip,
            width=\linewidth,
            height=\realsetupimageheight,
            keepaspectratio
        ]{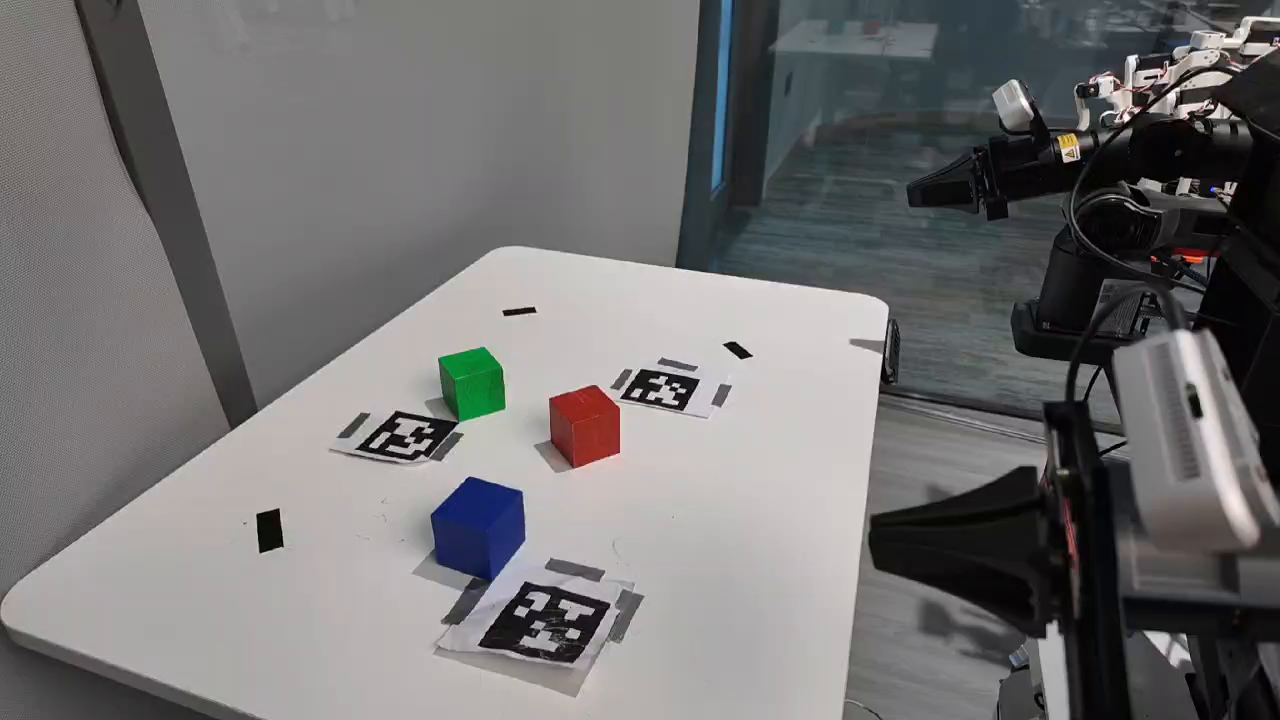}
        \caption{Stack Three Blocks}
        \label{fig:real_blocks}
    \end{subfigure}
    \caption{\textbf{Real-world platform and tasks.} (a) Bimanual platform
    (image: \href{https://www.agilex.ai/products/piper-dual-arm}{AgileX
    Robotics}). (b--d) Initial task configurations.}
    \label{fig:real_world_setup}
    \end{minipage}\hfill
    \begin{minipage}[c]{0.47\linewidth}
    \centering
    \captionsetup{type=table,position=top}
    \caption{\textbf{Real-world success counts} (20 trials per task) on the
    tasks in Figure~\ref{fig:real_world_setup}b--d. Shared WAM and
    Utility-WAM (Correction) follow the configurations of
    Table~\ref{tab:component_ablation} and are evaluated under the same protocol.
    Bold and underlined entries denote the best and second-best results in
    each column, respectively (including ties).}
    \label{tab:real_world_results}
    \small
    \setlength{\tabcolsep}{3pt}
    \begin{adjustbox}{max width=\linewidth}
    \begin{tabular}{lcccc}
        \toprule
        Method & Cup & Can & Blocks & Total \\
        \midrule
        FAST-WAM & \textbf{18/20} & 14/20 & 11/20 & 43/60 (71.7\%) \\
        Shared WAM (wo) & \underline{17/20} & 16/20 & 14/20 & 47/60 (78.3\%) \\
        Shared WAM (w) & 16/20 & 16/20 & 13/20 & 45/60 (75.0\%) \\
        Utility-WAM (Correction)
        & \textbf{18/20} & \underline{17/20} & \underline{15/20}
        & \underline{50/60 (83.3\%)} \\
        \rowcolor{oursblue}
        Utility-WAM (Selective)
        & \textbf{18/20} & \textbf{18/20} & \textbf{17/20}
        & \textbf{53/60 (88.3\%)} \\
        \bottomrule
    \end{tabular}
    \end{adjustbox}
    \end{minipage}
\end{figure}

\paragraph{Setup.}
Figure~\ref{fig:real_world_setup} shows the bimanual platform and task
configurations. We use two AgileX PiPER arms with three RGB views and
evaluate three tasks: \textit{Place Empty Cup}, \textit{Insert Can}, and
\textit{Stack Three Blocks}. Each task has 200 teleoperated demonstrations.
FAST-WAM and Utility-WAM (Selective) are trained on the same demonstrations,
with Utility-WAM following the three-stage procedure of Section~\ref{sec:method},
and each method is evaluated on 20 trials per task with varied initial
layouts. Hardware, task definitions, and data collection details are in
Appendices~\ref{sec:hardware_task} and~\ref{app:real_data_training}.

\paragraph{Results.}
Utility-WAM (Selective) succeeds in 53 of 60 trials versus 43 for FAST-WAM
(Table~\ref{tab:real_world_results}), a 16.7-point improvement with an
unadjusted two-sided 95\% Newcombe interval of $[2.3,30.4]$ points. The two
methods tie on \textit{Place Empty Cup}; the gains come from
\textit{Insert Can} ($+4$ successes), which requires aligning the can with a
recess, and \textit{Stack Three Blocks} ($+6$), the longest multi-stage task,
consistent with the simulation finding that generated futures help most on
long-horizon tasks. This evaluation demonstrates that selective visual reasoning transfers effectively to physical hardware under matching demonstration budgets.

\section{Related Work}
\label{sec:related_works}

\paragraph{World Action Models and Test-Time Futures.}
WAMs learn actions together with future visual prediction, either jointly
\citep{wu2024unleashing,cheang2024gr,cen2025worldvla,guo2024prediction,
zhu2025unified,bi2026motus,kim2026cosmos,ye2026world} or through predictive
representations and inverse dynamics
\citep{tian2025predictive,hu2024video,li2026causal}. Their deployment choices
differ: FAST-WAM omits future generation, GigaWorld-Policy makes it optional,
AHA-WAM reuses video context, Efficient-WAM reduces its cost, and AdaWAM
routes it using phase supervision
\citep{yuan2026fast,ye2026gigaworld,cai2026aha,li2026efficient,
tang2026dreaming}. We instead study the induced action change: we first repair
it, then reuse the adapter's paired outcomes, rather than phase supervision,
to decide where futures are generated.

\paragraph{Privileged Futures and Distillation.}
PFD~\citep{fang2026privileged} distills ground-truth-future corrections into
a current-only policy, sacrificing predictive context at inference. In contrast,
Utility-WAM preserves test-time visual imagination on long-horizon horizons while
using adaptive gating to avert negative transfer on precision subtasks.

\paragraph{Imperfect Predictions and Negative Transfer.}
Learned rollouts are useful but fallible: I2A learns to interpret them, while
STEVE and MBPO limit their use
\citep{racaniere2017imagination,buckman2018sample,janner2019trust}. Training on
ground-truth inputs but deploying on model predictions creates exposure bias
or covariate shift
\citep{bengio2015scheduled,ross2011reduction,ning2023input}; generated
intermediate reasoning can likewise hurt in robotics and language models
\citep{zhao2025cot,liu2025mind}. Following negative-transfer definitions
\citep{wang2019characterizing,zhang2023survey}, we treat the predicted future
as the additional source while holding parameters and data fixed. Feature
contamination~\citep{zhang2024feature} offers a related
representation-level account.

\section{Conclusion and Discussion}
\label{sec:conclusion}

Visual futures in World Action Models induce an action delta $\Delta a$ relative to future-free execution. While this predictive delta provides essential guidance on long-horizon tasks, it remains largely redundant on standard manipulations and induces severe failures on precision tasks---all while imposing significant inference latency on every decision. 

Utility-WAM resolves this tension by decoupling test-time visual imagination into targeted action correction and selective deployment. A residual adapter attached to a frozen shared WAM directly aligns predicted velocity shifts with ground-truth guidance, eliminating the majority of severe task drops while preserving long-horizon gains. A lightweight utility gate, trained on paired counterfactual outcomes, selectively bypasses future generation when no tangible action benefit is anticipated, reducing video generation by up to 75.5\% and latency by 38\%.

\paragraph{Generality.}
The phenomenon of future-induced negative transfer recurs across diverse WAM paradigms, including joint denoising and inverse dynamics. Our correction framework applies broadly to any architecture with dual action paths and a flow-matching action head: joint-denoising WAMs~\citep{ye2026world,kim2026cosmos} expose these paths naturally through attention masks, while inverse-dynamics models~\citep{li2026causal,feng2507vidar} can attach the adapter directly to the action head.

\paragraph{Limitations.}
Our evaluations use a common Stage-1 checkpoint due to the computational cost of pretraining 5B diffusion backbones; while cross-task consistency and threshold robustness are verified, variance across independently retrained backbones remains uncharacterized. Furthermore, the utility gate relies on an offline action-error proxy rather than closed-loop task returns, leaving room for future work to explore online reinforcement-learned deployment policies.

\bibliography{iclr2027_conference}
\bibliographystyle{iclr2027_conference}

\clearpage

\appendix

\section{Implementation Details}
\label{app:implementation_details}

\subsection{Detailed Three-Stage Training and Inference}
\label{app:detailed_training}

\paragraph{Shared-WAM masks.}
Let $C$, $F$, and $A$ denote current-frame, future-video, and action tokens,
with each relation interpreted as query-to-key access. Following FAST-WAM,
$M_{\mathrm{wo}}$ permits $C{\rightarrow}C$, $F{\rightarrow}\{C,F\}$, and
$A{\rightarrow}\{C,A\}$, while $M_{\mathrm{w}}$ additionally permits
$A{\rightarrow}F$. At future-free inference, $F$ is omitted and the
remaining tokens use the corresponding submask. Both training paths retain
the video objective.

\paragraph{Adapter sampling.}
For each adapter update, we sample one of the ten video denoising updates and
roll out the frozen video model through it to obtain post-update features
$F_k^{+}$. The ground-truth future is noised to the matched video level with
the same initial video noise, while the action noise level follows the action
flow-matching scheduler. The no-future, predicted-future, and
ground-truth-future paths receive the same noisy expert action. The adapter
uses $q_\tau^a$ as queries and $F_k^{+}$ as keys and values, followed by a
feed-forward network and a zero-initialized output projection.

\paragraph{Gate labels.}
Each gate-training input contributes one paired action sample from the two
frozen paths. They share Gaussian action noise, solver, and action schedule;
the future-conditioned path additionally draws video noise. The gate is
trained after adapter correction. Subtracting and summing the paired Euler
updates cancels the common initial noise and gives
Eq.~\eqref{eq:correction_accumulation}; the resulting $\Delta a$ supervises
Stage~3.

\paragraph{Inference update.}
Starting from Gaussian noise, we solve Eq.~\eqref{eq:inference_routing} with
the ten-step explicit-Euler solver. In
the future-conditioned path, each iteration updates the video latent, extracts
$F_k^{+}$, refines the action velocity, and then updates the action latent.
The future-free path skips video updates. The exact schedule is given below,
and Appendix~\ref{app:threshold_selection} specifies gate calibration.

\subsection{Base Model and Training Process}
\label{app:model_and_hyperparameters}

\paragraph{Base Model.} The Wan backbone~\citep{wan2025wan} models videos in a spatiotemporal latent space. A pretrained VAE encodes video frames into compact representations, and the video DiT performs conditional generation in this space. We retain the pretrained video weights and initialize the action DiT backbone by interpolating compatible video DiT parameters to its target dimensions. During training, the VAE remains frozen, and language embeddings are precomputed and cached.

\paragraph{Training Hyperparameters.} Table~\ref{tab:training_hyperparameters} summarizes the model, inference, and three-stage training settings. For gate training, we retain the checkpoint with the lowest validation loss and apply early stopping with a patience of three epochs and a minimum improvement of $10^{-4}$.

\begin{table}[htbp]
      \centering
      \caption{Model and training hyperparameters. Paired values denote LIBERO / RoboTwin~2.0. Global batch sizes include gradient accumulation.}
      \label{tab:training_hyperparameters}

      \small
      \setlength{\tabcolsep}{4pt}
      \renewcommand{\arraystretch}{1.12}

      \begin{adjustbox}{max width=\linewidth}
      \begin{tabular}{lccc}
          \toprule
          \multicolumn{4}{l}{\textit{Model and inference settings}} \\
          \midrule

          Video backbone
          & \multicolumn{3}{c}{Wan2.2-TI2V-5B} \\

          Action DiT depth $L_a$
          & \multicolumn{3}{c}{30} \\

          Action DiT hidden dimension $d_a$
          & \multicolumn{3}{c}{1024} \\

          Action horizon $H$
          & \multicolumn{3}{c}{32} \\

          Video frames per sample $T$
          & \multicolumn{3}{c}{9} \\

          Multi-view fusion
          & \multicolumn{3}{c}{Concatenation before VAE encoding} \\

          Adapter fusion
          & \multicolumn{3}{c}{Cross-attention (action queries, video keys/values)} \\

          Adapter feature dimension $d_{\mathrm{adapt}}$
          & \multicolumn{3}{c}{256 / 512} \\

          Adapter output projection initialization
          & \multicolumn{3}{c}{Zero} \\

          Gate feature fusion
          & \multicolumn{3}{c}{MLP} \\

          Action denoising steps $N_a$
          & \multicolumn{3}{c}{10} \\

          Future-video updates $N_v$
          & \multicolumn{3}{c}{$\{0,10\}$, selected by the gate} \\

          \midrule
          Training hyperparameter
          & \shortstack{Stage 1\\Shared WAM}
          & \shortstack{Stage 2\\Adapter}
          & \shortstack{Stage 3\\Gate} \\
          \midrule

          Optimizer
          & AdamW & AdamW & AdamW \\

          Learning rate
          & $10^{-4}$ & $10^{-4}$ & $10^{-4}$ \\

          Weight decay
          & $10^{-2}$ & $10^{-4}$ & $10^{-4}$ \\

          Adam $(\beta_1,\beta_2)$
          & $(0.9,0.95)$ & $(0.9,0.95)$ & $(0.9,0.999)$ \\

          Global batch size
          & 128 / 128 & 48 / 96 & 64 / 64 \\

          Training budget (LIBERO)
          & 10 epochs & $30{,}000$ steps & Up to 20 epochs \\

          Training budget (RoboTwin)
          & 5 epochs & $20{,}000$ steps & Up to 20 epochs \\

          Learning-rate schedule
          & Cosine & Cosine & Constant \\

          Warmup ratio
          & 5\% & 5\% & -- \\

          Gradient clipping norm
          & 1.0 & 1.0 & 1.0 \\

          Training precision
          & BF16 & BF16 & FP32 \\

          \bottomrule
      \end{tabular}
      \end{adjustbox}
\end{table}

\paragraph{Adapter Architecture.} The residual adapter uses a single cross-attention block, where projected action features attend to spatially pooled future-video features, followed by a residual feed-forward network and a zero-initialized output projection that predicts action-velocity corrections. For LIBERO, we use a hidden dimension of 256 and a $256 \rightarrow 512 \rightarrow 256$ feed-forward network, resulting in approximately 1.58M trainable parameters. For RoboTwin~2.0, we increase the hidden dimension to 512 and use a $512 \rightarrow 1024 \rightarrow 512$ feed-forward network, resulting in approximately 4.21M trainable parameters. The adapter parameters are shared across denoising steps and are used only in the future-conditioned branch.

\paragraph{Gate Architecture.} The routing gate encodes each current camera view using a shared lightweight residual CNN. Each view is pooled into a $4\times4$ grid of 128-dimensional visual tokens, augmented with spatial position and camera embeddings. Language and proprioceptive features are fused into a 128-dimensional query, which attends to the visual tokens through a single four-head cross-attention block. A residual feed-forward network and an MLP prediction head produce a scalar logit, converted by a sigmoid into $g_{\omega}(x)$. We use the same architecture on both benchmarks, with approximately 1.3M trainable parameters. The gate operates exclusively on current inputs and makes its routing decision before future generation.
\paragraph{Flow-matching schedule and sampling order.}
We follow the released FAST-WAM continuous flow scheduler rather than sampling a logit-normal inference grid. Let $S=1000$ denote the number of training timesteps. We use the original paper and released-checkpoint setting $s=5.0$ for both the video and action modalities; the later repository revision that changes the default action shift to $1.0$ is not used in our experiments. The scheduler maps $u\in[0,1]$ to the noise level
\begin{equation}
    \phi_s(u)=\frac{s\,u}{1+(s-1)u},
    \qquad \tau=\phi_s(u),
    \qquad t=S\tau.
    \label{eq:fastwam_schedule}
\end{equation}
During training, the video and action branches draw their own $u\sim\mathcal{U}(0,1)$ independently. During inference with $N=10$ steps, we set $u_i=1-i/N$, $\tau_i=\phi_s(u_i)$, and $\Delta_i=\tau_{i+1}-\tau_i$ for $i=0,\ldots,N-1$. FAST-WAM uses the explicit-Euler update
\begin{equation}
    z_{i+1}=z_i+\Delta_i f_{\theta}(z_i,S\tau_i;\mathcal{C}),
    \label{eq:fastwam_euler}
\end{equation}
where $f_{\theta}$ is the velocity predictor and $\mathcal{C}$ its conditioning. The no-video branch applies these ten updates only to the action latent. The future-conditioned branch pairs video and action updates by the common index $i$: it updates the video latent first, extracts $F_{i+1}^{+}$, applies the adapter, and then updates the action latent. During adapter training, we instead sample a video update $k$ uniformly from $\{1,\ldots,10\}$ and sample the action noise level independently. Classifier-free guidance uses scale $1.0$. Each sample contains 32 actions and nine video frames, including the current frame, so adjacent video frames correspond to four action steps.

\paragraph{Loss weights.}
For shared-WAM training, we set $\lambda_{\mathrm{wo}}=\lambda_{\mathrm{w}}=0.5$ in Eq.~\eqref{eq:stage1_loss}. For adapter training, we set the alignment weight to $\alpha=1.0$ and the excess-error weight to $\beta=0.5$ in Eq.~\eqref{eq:stage2_loss}. The relative improvement margin in Eq.~\eqref{eq:target_correction} is $\rho=0.05$. The scheduler weight in Eq.~\eqref{eq:stage2_loss} is the FAST-WAM training weight $w(\tau)\propto r(\tau)-\min_i r(\tau_i)$ with $r(\tau)=\exp[-2(\tau-\tfrac{1}{2})^2]$, normalized to mean one over the discrete training grid $\tau_i=\phi_s(1-i/S)$, $i=0,\ldots,S-1$.

\subsection{Simulation Benchmarks}
\label{app:sim_bench}

\paragraph{LIBERO.} We follow the FAST-WAM data and evaluation pipeline on the Spatial, Object, Goal, and Long suites of LIBERO~\citep{liu2023libero}. Each suite contains 500 demonstrations spanning ten tasks. The model receives a fixed external view and a wrist view, each resized to $224{\times}224$ and concatenated horizontally before VAE encoding. Actions are seven-dimensional and proprioceptive states are eight-dimensional; action and state normalization follows the FAST-WAM min--max preprocessing. A sample contains 32 actions and nine video frames. At evaluation, the policy samples one chunk, executes its first ten actions, and replans, for a total of 2,000 trials across the 40 tasks.

\paragraph{RoboTwin 2.0.} We follow the FAST-WAM multi-task setup on RoboTwin~2.0~\citep{chen2025robotwin}, using 2,500 demonstrations from clean scenes and 25,000 demonstrations with heavy scene randomization over more than 50 bimanual tasks. Inputs comprise a high camera and two wrist cameras, each resized to $240{\times}320$ and combined using the FAST-WAM three-camera layout. Actions and proprioceptive states are both 14-dimensional and use the released z-score statistics. Each sample again contains 32 actions and nine video frames. At evaluation, the policy samples one chunk, executes its first 24 actions, and replans. We report 100 trials per task under both clean and randomized conditions.

\subsection{Reported Result Sources}
\label{app:baseline_protocols}

\paragraph{LIBERO provenance.}
LingBot-VA~\citep{li2026causal}, Cosmos Policy~\citep{kim2026cosmos}, AdaWAM~\citep{tang2026dreaming}, the three FAST-WAM variants~\citep{yuan2026fast}, and PFD~\citep{fang2026privileged} are copied from their respective original papers. The cited $\pi_{0.5}$ paper does not report LIBERO; its values are taken from the authors' official OpenPI LIBERO release\footnote{\href{https://github.com/Physical-Intelligence/openpi/blob/35106c9bea0f84584e1a9b3a9ee1ae7aba070532/examples/libero/README.md}{Official $\pi_{0.5}$ LIBERO results.}}.

\paragraph{RoboTwin 2.0 provenance.}
LingBot-VA~\citep{li2026causal}, Efficient-WAM~\citep{li2026efficient}, AHA-WAM~\citep{cai2026aha}, AdaWAM~\citep{tang2026dreaming}, the three FAST-WAM variants~\citep{yuan2026fast}, and PFD~\citep{fang2026privileged} are taken from their latest audited original-paper tables; PFD's Clean and Random scores (93.11 and 92.69) are rounded to one decimal place. The cited $\pi_{0.5}$ paper does not contain RoboTwin experiments; its values are traceable to LingBot-VA's Appendix comparison table~\citep{li2026causal} and are marked with a dagger. Utility-WAM values are produced by our evaluation protocol. Published averages are retained when available; otherwise, we compute the arithmetic mean of the displayed Clean and Random scores before rounding to one decimal place.

Tables~\ref{tab:component_ablation} and~\ref{tab:adapter_loss_ablation} provide the controlled comparisons for method attribution.

\section{Gate Threshold Selection}
\label{app:threshold_selection}

\paragraph{Budgeted routing objective.}
Under a budget $B$ on the fraction of queries that generate video
(equivalently, $N_{\mathrm{eff}}=10\,\mathbb{E}_{x}[b(x)]$ video steps per
query), the gate of Section~\ref{sec:utility_gating} routes queries toward
corrections that are more likely to reduce action error:
\begin{equation}
    \max_{b:\mathcal{X}\rightarrow\{0,1\}}\;
    \mathbb{E}_{x}\!\left[b(x)\bigl(2y(x)-1\bigr)\right]
    \quad\text{s.t.}\quad
    \mathbb{E}_{x}\!\left[b(x)\right]\le B.
    \label{eq:budgeted_routing}
\end{equation}
With $p(x)=P(y=1\mid x)$, Eq.~\eqref{eq:budgeted_routing} is solved by
routing the queries with the largest $p(x)$ among those with $p(x)>1/2$, up to
budget $B$, so a score that ranks queries by $p(x)$ suffices.

\paragraph{Validation labels.}
We determine the routing threshold $\eta$ exclusively from the gate validation set, which is disjoint from the gate-training inputs and from all test episodes, with the WAM, adapter, and gate frozen. For each validation input $x_i$, we draw one paired future-conditioned/no-video action sample with the shared action noise of Section~\ref{sec:utility_gating}, compute $u_{\mathrm{act},i}$ by Eq.~\eqref{eq:future_utility}, set $y_i=\mathbf{1}[u_{\mathrm{act},i}>0]$, and record the gate logit $h_i=\log[g_{\omega}(x_i)/(1-g_{\omega}(x_i))]$. Each validation input thus contributes exactly one label--logit pair; no test data are used.

\paragraph{Hard budget.}
Because binary cross-entropy is a proper scoring rule for $y$, ranking
$g_{\omega}(x)$ approximates ranking the conditional probability of a
beneficial branch correction. For a prescribed budget $B$, we use the validation
quantile
\begin{equation}
    \eta_B =
    \max\!\left\{
        \frac{1}{2},
        Q_{1-B}\!\left(
            \{g_{\omega}(x_i)\}_{x_i\in\mathcal{D}_{\mathrm{val}}}
        \right)
    \right\},
    \qquad
    b_{\omega,B}(x)=
    \mathbf{1}\!\left[g_{\omega}(x)>\eta_B\right],
    \label{eq:budget_quantile}
\end{equation}
which bounds the empirical validation routing rate by $B$ (up to ties) while
excluding scores that do not predict a beneficial correction.

\paragraph{Default operating point.}
Without an external budget, we fit Gaussians with a shared variance to the
logits of the two classes, $h\mid y=j\sim\mathcal{N}(\mu_j,\sigma^2)$ with
$\mu_j$ the class-mean logit, and apply the Bayes rule under equal priors and
misclassification costs. Assuming $\mu_1>\mu_0$, the decision boundary is the
midpoint of the class means, independent of $\sigma^2$, so
\begin{equation}
    \eta=
    \left[
        1+\exp\!\left(-\frac{\mu_0+\mu_1}{2}\right)
    \right]^{-1}.
    \label{eq:calibrated_gate_threshold}
\end{equation}
This rule does not require calibrated gate probabilities. Calibration uses 5,408 LIBERO and 10,892 RoboTwin~2.0 validation inputs. The fitted logit boundaries are $h^{*}=-0.2452$ and $h^{*}=0.1644$, which yield $\eta=0.439$ and $\eta=0.541$, respectively, and each threshold is kept fixed throughout test evaluation. A deployment with a hard latency budget instead uses Eq.~\eqref{eq:budget_quantile}, moving along the frontier in Figure~\ref{fig:neff_success_curves}.

\section{Simulation Setup and Additional Results}

\subsection{Datasets and Evaluation Protocols}
\label{app:sim_protocols}

\paragraph{Validation Set and Threshold Selection.} We reserve a task-stratified validation set whose trajectories are disjoint from those used to train the gate. For LIBERO, this set contains 169 trajectories, with 32 temporally distributed observation points per trajectory, yielding 5,408 samples. For RoboTwin~2.0, it contains 2,723 trajectories, with four observation points per trajectory, yielding 10,892 samples. Appendix~\ref{app:threshold_selection} describes how the routing threshold is determined from this set; no test rollouts are used.

\paragraph{Closed-Loop Evaluation.} Each policy query predicts a chunk of 32 actions using 10 action denoising steps. The policy executes the first 10 actions on LIBERO and the first 24 actions on RoboTwin~2.0 before obtaining a new observation and replanning. At each replanning step, the gate applies the fixed benchmark-specific threshold to determine whether to generate future observations. Task success is determined by the benchmark-provided success checks. Each rollout terminates upon success or upon reaching the task-dependent execution limit in the FAST-WAM evaluation implementation. All configurations are evaluated on the same episodes: for each task and condition, the $i$-th trial uses the same initial state (object layout and scene randomization) and the same random seeds under every configuration, so all between-configuration comparisons are paired at the episode level. We report success rates measured over the closed-loop evaluation rollouts.

\subsection{Replication Across Future-Usage Mechanisms}
\label{app:replication}

Table~\ref{tab:replication} compares our shared-checkpoint sign reversal with
the two future-generating variants of FAST-WAM~\citep{yuan2026fast}: Joint,
which jointly denoises future video and actions, and IDM, which generates a
future before an inverse-dynamics model. In all three comparisons, future
conditioning raises the LIBERO average, with the largest gain on Long, and
lowers RoboTwin~2.0 success under both clean and randomized conditions. The
FAST-WAM comparisons are not controlled: each future-generating variant is
trained separately, so they confound future conditioning with training
differences. We use them only to show that the sign reversal is not specific
to our checkpoint and rely on the shared-checkpoint comparison for
attribution.

\begin{table}[htbp]
    \centering
    \caption{\textbf{Replication of the benchmark-level sign reversal.}
    Entries are $\Delta J$, future-conditioned minus future-free success in
    percentage points; negative values indicate harm. FAST-WAM rows use the published results of
    \citet{yuan2026fast}, comparing separately trained models; the last row is
    the shared-checkpoint comparison of Table~\ref{tab:component_ablation}.}
    \label{tab:replication}
    \small
    \setlength{\tabcolsep}{4pt}
    \begin{adjustbox}{max width=\linewidth}
    \begin{tabular}{llccccc}
        \toprule
        & & \multicolumn{2}{c}{LIBERO} & \multicolumn{3}{c}{RoboTwin~2.0} \\
        \cmidrule(lr){3-4}\cmidrule(lr){5-7}
        Comparison & Stage 1 & Avg. & Long & Clean & Random & Avg. \\
        \midrule
        FAST-WAM-Joint $-$ FAST-WAM & Separate & $+0.9$ & $+1.6$ & $-1.0$ & $-1.5$ & $-1.3$ \\
        FAST-WAM-IDM $-$ FAST-WAM   & Separate & $+0.4$ & $+2.4$ & $-0.7$ & $-0.4$ & $-0.6$ \\
        Shared WAM (w) $-$ Shared WAM (wo) & Shared & $+0.8$ & $+2.4$ & $-1.1$ & $-0.5$ & $-0.8$ \\
        \bottomrule
    \end{tabular}
    \end{adjustbox}
\end{table}

Using the published per-task RoboTwin~2.0 results of FAST-WAM, we compute
$\Delta J_{t}$ for each of the 50 tasks, averaged over clean and
randomized conditions (Table~\ref{tab:replication_fastwam_tasks}). For both
variants, future generation helps on more tasks than it hurts, and the median
task change is non-negative; the aggregate is nonetheless negative because a few
tasks fail severely. The task-level changes of the two variants are strongly
rank-correlated (Spearman $0.84$), and \emph{Press Stapler} and \emph{Place Can
Basket} are among the three most harmed tasks for both. Changes estimated
separately under clean and randomized conditions, which share no evaluation
episodes, are also positively correlated (Spearman $0.42$ for Joint and
$0.40$ for IDM). These
statistics are descriptive, but they indicate that harm recurs on a subset of
tasks rather than spreading uniformly, so future use should be corrected
where harm occurs rather than enabled or disabled globally.

\begin{table}[htbp]
    \centering
    \caption{\textbf{Task-level negative transfer in FAST-WAM on
    RoboTwin~2.0.} $\Delta J_{t}$ is the future-generating variant minus
    FAST-WAM, averaged over clean and randomized success, computed from the
    published per-task results of \citet{yuan2026fast}. Negative values
    indicate harm.}
    \label{tab:replication_fastwam_tasks}
    \small
    \setlength{\tabcolsep}{4pt}
    \begin{adjustbox}{max width=\linewidth}
    \begin{tabular}{lcccccl}
        \toprule
        Variant & Harmed & Helped & Tied & Median
        & $\geq 5$ harmed / helped & Most harmed task \\
        \midrule
        FAST-WAM-Joint & 12/50 & 26/50 & 12/50 & $+0.5$ & 6 / 5
        & Open Microwave ($-45.0$) \\
        FAST-WAM-IDM   & 17/50 & 20/50 & 13/50 & $0.0$ & 5 / 8
        & Press Stapler ($-40.0$) \\
        \bottomrule
    \end{tabular}
    \end{adjustbox}
\end{table}

\subsection{Task-Level Negative-Transfer Analysis}
\label{app:task_level_ntg}

\paragraph{Metrics.}
For task $t$, we compute
$\Delta J_{t}(\pi)=J_{t}(\pi)-J_{t}(\pi_{\mathrm{wo}})$ against the
future-free branch of the same Stage-1 checkpoint. We report the number of
tasks with $\Delta J_{t}(\pi_{\mathrm{w}})<0$, the median across tasks, and
the range. The sign count is a descriptive prevalence statistic: with finite
rollouts, a negative point estimate is not by itself evidence that an
individual task has nonzero population $\Delta J_{t}$.

\paragraph{Per-task changes on RoboTwin~2.0.}
Table~\ref{tab:robotwin_per_task_ntg} lists $\Delta J_{t}$ for the ten
most harmed RoboTwin~2.0 tasks, alongside FAST-WAM's separately trained Joint
and IDM variants, and summarizes all 50 tasks. Our three columns share the
future-free reference, so the change from the always-video column to the
correction and selective columns directly measures how much of the loss each
component removes.

\paragraph{Why the per-task columns are highly consistent.}
All configurations are evaluated on identical initial states and random seeds
(Appendix~\ref{app:sim_protocols}), so each entry is a paired difference over
the same 200 episodes and a step of 0.5 points is a single episode. Because
the adapter and the gate leave the Shared WAM frozen, most episodes end with
the same outcome under every configuration: across all 50 tasks, adjacent
columns differ by at most four episodes (2.0 points) except on the two severe
tasks. The paired design removes variation in initial states and sampling
noise, but not training-seed variation, so small per-task differences should
be read as descriptive trends.

\begin{table}[htbp]
    \centering
    \caption{\textbf{Per-task success changes on RoboTwin~2.0: the ten
    most harmed tasks.} All entries are $\Delta J_{t}$ in percentage
    points relative to the future-free reference, averaged over clean
    and randomized conditions; negative values indicate harm. Tasks are ranked
    by the mean $\Delta J_{t}$ over the three future-conditioned comparisons
    (Joint, IDM, and Shared WAM (w)), most negative first, with ties broken by
    Shared WAM (w). FAST-WAM columns use the published per-task results of
    \citet{yuan2026fast} with FAST-WAM as reference. Our columns use Shared
    WAM (wo) as the common reference and are paired over the same episodes.
    The bottom rows summarize all 50 tasks.}
    \label{tab:robotwin_per_task_ntg}
    \small
    \setlength{\tabcolsep}{4pt}
    \begin{tabular}{lccccc}
      \toprule
      & \multicolumn{2}{c}{FAST-WAM}
      & \multicolumn{3}{c}{Ours} \\
      \cmidrule(lr){2-3}\cmidrule(lr){4-6}
      Task
      & Joint & IDM
      & Shared WAM (w) & Correction & Selective \\
      \midrule
      Press Stapler       & $-42.5$ & $-40.0$ & $-54.0$ & $-6.0$  & $-6.5$ \\
      Place Can Basket    & $-33.5$ & $-37.5$ & $-34.0$ & $-14.0$ & $-17.0$ \\
      Open Microwave      & $-45.0$ & $0.0$   & $-1.5$  & $-1.0$  & $-0.5$ \\
      Blocks Ranking Size & $-9.0$  & $-11.5$ & $-2.5$  & $-2.0$  & $-0.5$ \\
      Open Laptop         & $-8.5$  & $-7.0$  & $-2.5$  & $-2.0$  & $-0.5$ \\
      Place Object Basket & $-5.0$  & $-4.0$  & $0.0$   & $0.0$   & $0.0$ \\
      Place Dual Shoes    & $0.0$   & $-5.0$  & $-0.5$  & $-0.5$  & $-0.5$ \\
      Adjust Bottle       & $-1.5$  & $-3.5$  & $0.0$   & $0.0$   & $0.0$ \\
      Place Shoe          & $-1.5$  & $-1.0$  & $-1.0$  & $-0.5$  & $0.0$ \\
      Rotate QRcode       & $+0.5$  & $-2.5$  & $-1.0$  & $-0.5$  & $0.0$ \\
      \midrule
      Harmed tasks (all 50)
      & 12/50 & 17/50 & 21/50 & 18/50 & 10/50 \\
      Median (all 50)
      & $+0.5$ & $0.0$ & $0.0$ & $0.0$ & $0.0$ \\
      \bottomrule
  \end{tabular}
\end{table}
  
\paragraph{Does correction target the harmed tasks?}
For the always-future adapter, define the task-level recovery
$R_{t}=J_{t}(\pi_{\mathrm{w+adapter}})
-J_{t}(\pi_{\mathrm{w}})$. We compare mean recovery for harmed
($\Delta J_{t}(\pi_{\mathrm{w}})<0$) and non-harmed tasks, that is, whether correction
preferentially repairs tasks harmed by predicted futures or shifts all tasks
uniformly.

\begin{table}[htbp]
    \centering
    \caption{\textbf{Adapter recovery on harmed and non-harmed tasks.}
    Harmed tasks have $\Delta J_{t}<0$ under Shared WAM (w).
    Recovery values are percentage points; positive values indicate
    improvement.}
    \label{tab:ntg_adapter_association}
    
    \small
    \setlength{\tabcolsep}{4pt}
    \begin{adjustbox}{max width=\linewidth}
    \begin{tabular}{lcc}
        \toprule
        Benchmark & Recovery, harmed & Recovery, non-harmed \\
        \midrule
        LIBERO & $+1.43$ & $+0.36$ \\
        RoboTwin~2.0 & $+3.55$ & $+0.36$ \\
        \bottomrule
    \end{tabular}
    \end{adjustbox}
\end{table}

On both benchmarks, harmed tasks recover more than non-harmed tasks. On
RoboTwin~2.0 the difference comes from the two severe tasks; the other 19
harmed tasks lose at most 2.5 points and recover $+0.34$ points on average,
matching the $+0.36$ on non-harmed tasks. Correction
therefore repairs severe harm where it occurs and otherwise acts as a small
uniform gain, consistent with Section~\ref{sec:mitigation}. The harmed and
non-harmed means are computed on all trials and may be inflated by regression
to the mean; the externally selected tasks below avoid this selection.

\paragraph{Externally identified harmed tasks.}
Selecting tasks by their observed $\Delta J_{t}$ and then measuring recovery
on the same trials invites regression to the mean. As a complementary check,
we examine the two severe tasks, which FAST-WAM's independently trained and
evaluated per-task results already identify as severely harmed
(Table~\ref{tab:replication_fastwam_tasks}), so their recovery is not
inflated by selecting on our own evaluation noise. The adapter removes 89\%
and 59\% of their losses (Table~\ref{tab:robotwin_per_task_ntg}). The gate
plays a complementary role: it improves benchmark-level success at lower cost
(Section~\ref{sec:mitigation}), while on these two tasks the residual losses
change from 6.0 to 6.5 and from 14.0 to 17.0 points. Their recovery therefore
comes from correction, consistent with the gate selecting between complete
action paths rather than repairing the future-conditioned path. The gate's
gain instead comes from the remaining 48 RoboTwin~2.0 tasks, where correction
changes little: relative to Shared WAM (wo), they gain 1.0 point under
uncorrected always-video inference, 1.35 under always-on correction, and about
1.9 under selective deployment, despite far less video generation (see
Table~\ref{tab:adapter_loss_ablation} for the complete ablation breakdown).

\subsection{Inference Efficiency}
\label{app:inference_efficiency}

\begin{wraptable}{r}{0.62\linewidth}
      \centering
      \setlength{\belowcaptionskip}{4pt}
      \caption{Mean video generation steps and inference
      latency per action query. Bold and underlined values denote the
      best and second-best results, respectively (lower is better).}
      \label{tab:inference_efficiency}

      \small
      \setlength{\tabcolsep}{3pt}
      \renewcommand{\arraystretch}{1.08}

      \begin{adjustbox}{max width=\linewidth}
      \begin{tabular}{@{}lcccc@{}}
          \toprule
          & \multicolumn{2}{c}{LIBERO}
          & \multicolumn{2}{c}{RoboTwin 2.0} \\
          \cmidrule(lr){2-3}
          \cmidrule(l){4-5}

          Method
          & $N_{\mathrm{eff}}$ & \shortstack{Latency\\(ms)$\downarrow$}
          & $N_{\mathrm{eff}}$ & \shortstack{Latency\\(ms)$\downarrow$} \\
          \midrule

          FAST-WAM
          & \textbf{0.00} & \underline{275.5}
          & \textbf{0.00} & \textbf{279.0} \\

          Shared WAM (wo-video)
          & \textbf{0.00} & \textbf{274.4}
          & \textbf{0.00} & \underline{280.0} \\

          Shared WAM (w-video)
          & 10.00 & 467.5
          & 10.00 & 528.7 \\

          \textbf{Utility-WAM}
          & \underline{8.70} & 436.3
          & \underline{2.45} & 328.2 \\
          \bottomrule
      \end{tabular}
      \end{adjustbox}
  \end{wraptable}

We measure end-to-end inference latency from CPU inputs to CPU action outputs on a single H100 80GB GPU, using BF16 and a batch size of 1. Table~\ref{tab:inference_efficiency} reports latency and $N_{\mathrm{eff}}$. Compared with the shared WAM that always generates future video, Utility-WAM reduces latency by 6.7\% on LIBERO and 37.9\% on RoboTwin~2.0. Future-free inference remains faster: relative to Shared WAM (wo-video), Utility-WAM adds 59\% latency on LIBERO for $+1.5$ points and 17\% on RoboTwin~2.0 for $+1.4$ points, compared with 70\% and 89\% for always generating futures, which gains $+0.8$ points on LIBERO and loses $0.8$ points on RoboTwin~2.0.

\section{Real-World Experimental Details and Qualitative Results}

\subsection{Real-World Hardware and Task Details}
\label{sec:hardware_task}

\paragraph{Hardware Platform.}
As shown in Figure~\ref{fig:real_world_setup}, our real-world platform consists of two six-DoF AgileX PiPER manipulators, each equipped with a two-finger gripper. The observation system comprises one fixed front-view RGB camera and one wrist-mounted RGB camera on each arm. The robot state is represented by a 14-dimensional vector containing the six joint positions and gripper opening of each arm.

\paragraph{Visual Observations.}
The recorded images for \textit{Place Empty Cup} and \textit{Stack Three Blocks} have a resolution of $640 \times 480$ pixels for all three views. For \textit{Insert Can}, the recorded front-view images have a resolution of $848 \times 480$ pixels, while both wrist views have a resolution of $640 \times 480$ pixels. All resolutions are reported as width $\times$ height. In the training pipeline, each view is resized to $320 \times 240$ pixels, and the three views are combined into a multi-view image of $320 \times 384$ pixels.

\paragraph{Action Representation and Execution.}
The policy uses a 14-dimensional absolute-position action representation comprising the joint and gripper targets of both arms. The training data are sampled at 10\,Hz, and each training window contains 32 action steps, corresponding to a 3.2\,s horizon.

\paragraph{Workspace and Initialization.}
All tasks are performed on a tabletop within the reachable workspace of the bimanual platform. The manipulated objects and their targets are placed in task-relevant regions that permit the required grasping, transport, placement, and insertion motions. Initial object placements are varied within these regions across trials while preserving task feasibility and the intended object--target relationships.

\paragraph{Tasks and Success Criteria.}
The initial task configurations are illustrated in Figures~\ref{fig:real_cup}--\ref{fig:real_blocks}. For \textit{Place Empty Cup}, the robot must grasp an empty cup and place it upright on the designated coaster. Success requires the cup to remain stably supported by the coaster after gripper release. For \textit{Insert Can}, the robot must grasp a can and insert it into the designated recess in a case. Success requires the can to be seated in the recess and remain upright after release; a can resting on the rim does not satisfy the task objective. For \textit{Stack Three Blocks}, the robot must sequentially arrange three blocks into a single tower. Success requires the completed tower to remain stable without gripper support.

\subsection{Demonstration Collection}
\label{app:real_data_training}

\paragraph{Demonstration Collection.}
We collect 200 demonstrations per task using AgileX's official teleoperation setup\footnote{\url{https://github.com/agilexrobotics/openpi-agilex}}. During teleoperation, we record RGB observations from the fixed front-view camera and both wrist-mounted cameras, together with the measured joint positions and gripper openings of both arms. Each demonstration is associated with a task instruction. The recorded robot state is a 14-dimensional vector containing six joint positions and one gripper opening per arm.

\subsection{Evaluation Protocol and Representative Rollouts}
\label{app:real_world_rollouts}

\paragraph{Evaluation protocol.}
Each configuration in Table~\ref{tab:real_world_results} is evaluated on 20
trials per task with varied initial layouts. Success requires stable placement on the coaster,
a seated upright can, or a stable three-block tower, respectively.
Table~\ref{tab:real_world_results} reports the per-task success counts.

\begin{figure*}[ht]
    \centering
    \includegraphics[width=\textwidth]{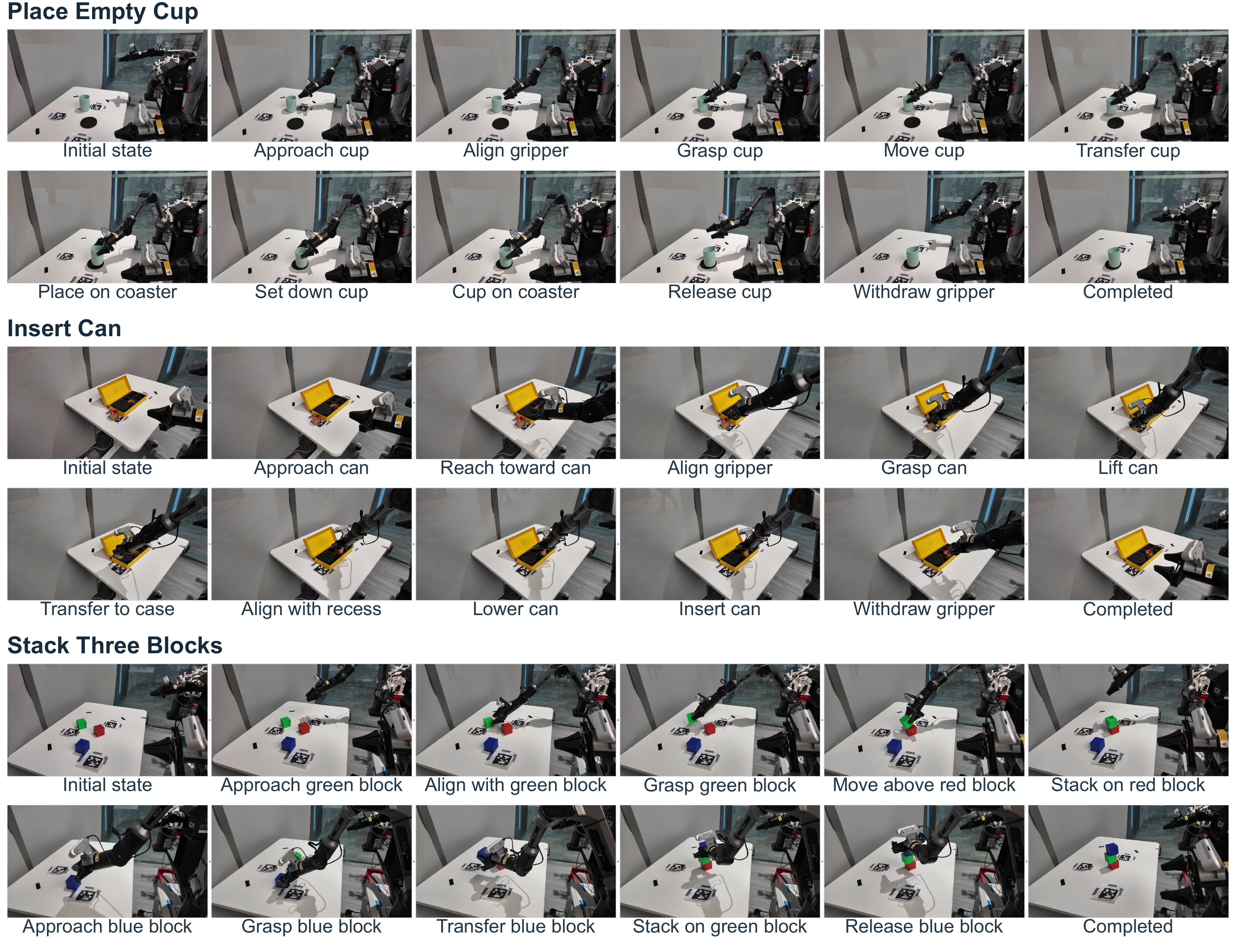}
    \caption{\textbf{Representative successful real-world evaluation sequences.}
    Keyframes for \textit{Place Empty Cup}, \textit{Insert Can}, and
    \textit{Stack Three Blocks} are shown from top to bottom and progress
    from left to right, illustrating the main manipulation stages and the
    final task configurations after gripper withdrawal. These examples are
    qualitative and are not used for method attribution.}
    \label{fig:real_world_timelines}
\end{figure*}

Figure~\ref{fig:real_world_timelines} presents keyframe sequences extracted from successful real-world evaluation videos of Utility-WAM. For \textit{Place Empty Cup}, the robot grasps the cup, transfers it to the designated coaster, and releases it in an upright position. For \textit{Insert Can}, the robot picks up the can, moves it toward the case, aligns it with the target recess, and completes the insertion before withdrawing the gripper. For \textit{Stack Three Blocks}, the robot first places the green block on the red block and then adds the blue block on top, forming a stable three-block tower. The final frame of each sequence shows the completed task configuration after gripper withdrawal. These sequences illustrate successful executions but are not comparative evidence.

\end{document}

%% file: math_commands.tex
\usepackage{amsmath,amsfonts,bm}

\def\1{\bm{1}}

\DeclareMathAlphabet{\mathsfit}{\encodingdefault}{\sfdefault}{m}{sl}
\SetMathAlphabet{\mathsfit}{bold}{\encodingdefault}{\sfdefault}{bx}{n}

